\documentclass[10pt,twocolumn,letterpaper]{article}

\usepackage{iccv}
\usepackage{times}
\usepackage{epsfig}
\usepackage{graphicx}
\usepackage{amsmath}
\newtheorem{definition}{Definition} 
\newtheorem{theorem}{Theorem}
\usepackage{amssymb}
\usepackage{multirow}
\usepackage{booktabs}
\usepackage{algorithm}
\usepackage{color}
\usepackage{marvosym}
\usepackage[accsupp]{axessibility}  
\usepackage{algorithmic}

\usepackage{enumitem}
\newcommand{\hut}{\textcolor{black}}
\newcommand{\yi}{\textcolor{black}}
\newcommand{\hutnew}{\textcolor{black}}
\newcommand{\leone}{\textcolor{black}}

\usepackage[pagebackref=true,breaklinks=true,letterpaper=true,colorlinks,bookmarks=false]{hyperref}

\iccvfinalcopy 

\def\iccvPaperID{3844} 

\ificcvfinal\fi

\begin{document}

\title{Phasic Content Fusing Diffusion Model with Directional Distribution Consistency for Few-Shot Model Adaption}


\author{Teng Hu$^{1}$\footnotemark[1],~~~Jiangning Zhang$^{2}$\thanks{Equal contributions.},~~~Liang Liu$^{2}$,~~~Ran Yi$^{1}$\thanks{Corresponding author.},~~~Siqi Kou$^{1}$\\Haokun Zhu$^{1}$, ~~~Xu Chen$^{2}$,~~~Yabiao Wang$^{2,3}$,~~~Chengjie Wang$^{1,2}$,~~~Lizhuang Ma$^{1}$
\\ 
$^1$Shanghai Jiao Tong University~~~$^2$Youtu Lab, Tencent~~~$^3$Zhejiang University\\
\tt\small $\{$hu-teng, ranyi, happy-karry, zhuhaokun, ma-lz$\}$@sjtu.edu.cn;\\
\tt\small $\{$vtzhang, leoneliu, cxxuchen, caseywang, jasoncjwang$\}$@tencent.com;\\
}

\maketitle
\ificcvfinal\thispagestyle{empty}\fi

\begin{abstract}
Training a generative model with limited number of samples is a challenging task. Current methods primarily rely on few-shot model adaption to train the network. However, in scenarios where data is extremely limited (less than 10), the generative network tends to overfit and suffers from content degradation.
To address these problems, we propose 
\yi{a novel phasic content fusing few-shot diffusion model with directional distribution consistency loss,}
which targets different learning objectives at distinct training stages of the diffusion model. 
\hutnew{Specifically, we design a phasic training strategy with phasic content fusion to help our model learn content and style information when t is large, and learn local details of target domain when t is small, leading to an improvement in the capture of content, style and local details.}
Furthermore, we introduce a novel directional distribution consistency loss that ensures the consistency between the generated and source distributions more efficiently and stably than the prior methods, \hutnew{preventing our model from overfitting}. Finally, we \hutnew{propose} a cross-domain structure guidance strategy that enhances structure consistency \hutnew{during domain adaptation.}
Theoretical analysis, qualitative and quantitative experiments demonstrate the superiority of our approach in few-shot generative model adaption tasks compared to state-of-the-art methods. The source code is available at: \href{https://github.com/sjtuplayer/few-shot-diffusion}{https://github.com/sjtuplayer/few-shot-diffusion}.
\end{abstract}

\section{Introduction}

\begin{figure}[t]
\centering
\includegraphics[width=0.47\textwidth]{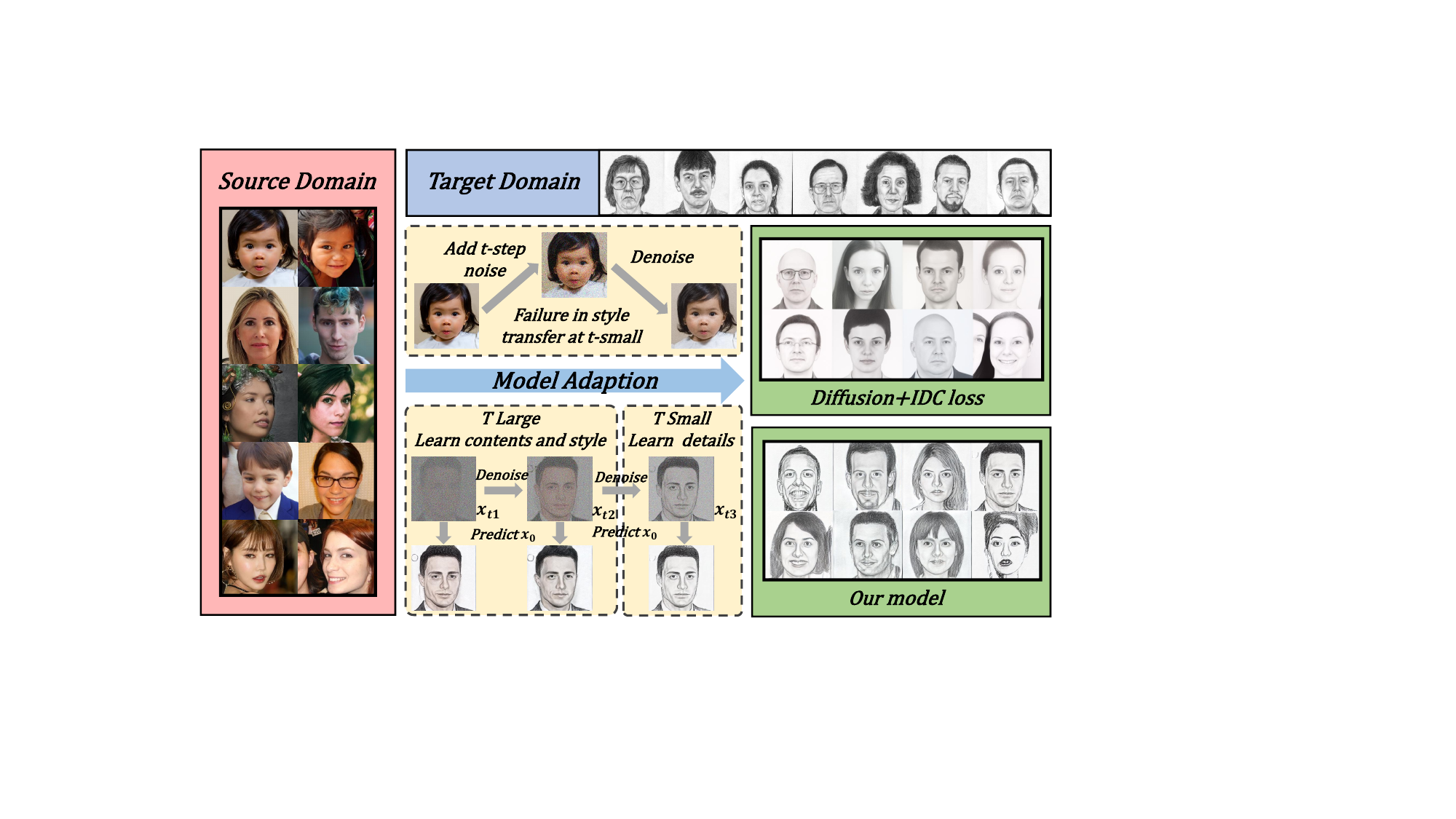}
\caption{Comparison with the diffusion model~\cite{zhu2022few} directly trained with IDC loss~\cite{ojha2021few}, which captures an inaccurate style due to the failed style transfer when $t$ is small.}
\label{fig:figure 1}
\vspace{-0.15in}
\end{figure}
Deep generative models~\cite{goodfellow2020generative,ho2020denoising} have achieved significant success in image generation tasks in recent years~\cite{zhang2021real,xu2022designing}. However, when the number of samples is limited, i.e., under few-shot image generation, they still suffer from the problem of overfitting. 
Most of the few-shot generative models are based on Generative Adversarial Networks (GANs)~\cite{goodfellow2020generative, bartunov2018few, clouatre2019figr, liang2020dawson, wang2018transferring} using few-shot model adaption. Some existing works have attempted to mitigate the overfitting problem through regularization or data augmentation~\cite{li2020few, zhang2019consistency, tran2021data, zhao2020differentiable, zhao2020image}, but still face difficulties when the samples are extremely limited (less than 10).
Recently, IDC~\cite{ojha2021few} and RSSA~\cite{xiao2022few} propose new cross-domain consistency loss functions to maintain similarity between the generated and original distributions and demonstrate promising results. However, due to the inherent limitations of GAN's architecture and generation process, there is still room for improvement for these methods in terms of preserving content information and enhancing image quality.

Over the last few years, diffusion models~\cite{ho2020denoising} have shown great success in image generation and have surpassed GAN model in sub-tasks like text-to-image synthesis and image inpainting~\cite{rombach2022high}.
\yi{Especially, the} flexible controlling process and \yi{good} generation quality \yi{of diffusion models can help} enhance the content information and structure consistency during domain adaption \yi{and are suitable for few-shot image generation task, which inspires us to study few-shot diffusion generation}. However, training few-shot diffusion model faces the following problems: (1) diffusion model tends to overfit with limited number of samples as GANs do; \hut{(2) simply training diffusion model with the few-shot loss functions in GAN \cite{ojha2021few,xiao2022few} leads to \yi{failed} style transfer at the detail learning stage ($t$ small), causing unsuccessful style capture as Fig.~\ref{fig:figure 1} shows}; \hut{(3) the existing loss in few-shot GAN adaptation only constrains the pairwise \yi{distances of generated samples in target and source domains to be similar,} 
leading to distribution rotation during training process, which may cause unstable and ineffective training.}
  
 To solve these problems, 
\leone{we propose a novel few-shot diffusion model that incorporates a phasic content fusing module and a directional distribution consistency loss to prevent overfitting and maintain content consistency.}
\hut{Specifically, we first design a phasic training strategy with phasic content fusion module, which integrates content information into the network and explicitly decomposes the model training into two stages: }learn content and style information \yi{when} $t$ is large, and learn local details in the target domain \yi{when} $t$ is small, 
\hut{preventing our model from confusion between content and target-domain local details effectively.}
Then, with a deep analysis on existing few-shot losses~\cite{ojha2021few,xiao2022few}, we propose a novel directional distribution consistency loss which \hut{can avoid the distribution rotation problem \yi{during training} and better keep the structure of generated distribution, improving the training stability, efficiency and solving the overfitting problem.}
\leone{Finally, we design a cross-domain structure guidance strategy to further integrate structural information during inference time, resulting in improved performance in both structure preservation and domain adaptation.}

Extensive qualitative and quantitative experiments show that our model outperforms the state-of-the-art few-shot generative models in both content preservation and domain adaptation. Moreover, through theoretical analysis, we also prove the effectiveness of our directional distribution consistency loss and the cross-domain structure guidance strategy in terms of distribution and \yi{structure} maintenance.

Our contributions can be summarized into three aspects:

\begin{itemize}
    \item
    \hut{We propose a novel phasic content fusing few-shot diffusion model, which learns content and style information \yi{when} $t$ is large, and learns local details \yi{when} $t$ is small. By incorporating the phasic content fusion module, our model excels in both content preservation and domain adaptation.}
    \item
    We design a \yi{novel} directional distribution consistency loss, \yi{which can effectively avoid the distribution rotation problem during training and better keep the structure of generated distribution. It} has been theoretically and experimentally proved \yi{that the directional distribution consistency loss can} maintain the structure of generated distribution in a more effective and stable way than the state-of-the-art methods.
    \item
    An iterative cross-domain structure guidance strategy is proposed to 
    \yi{further integrate structural information during inference time,} and has been demonstrated to achieve superior structure preserving performance in domain translation.
\end{itemize}

\section{Related Works}
\begin{figure*}[t]
\centering
\includegraphics[width=0.83\textwidth]{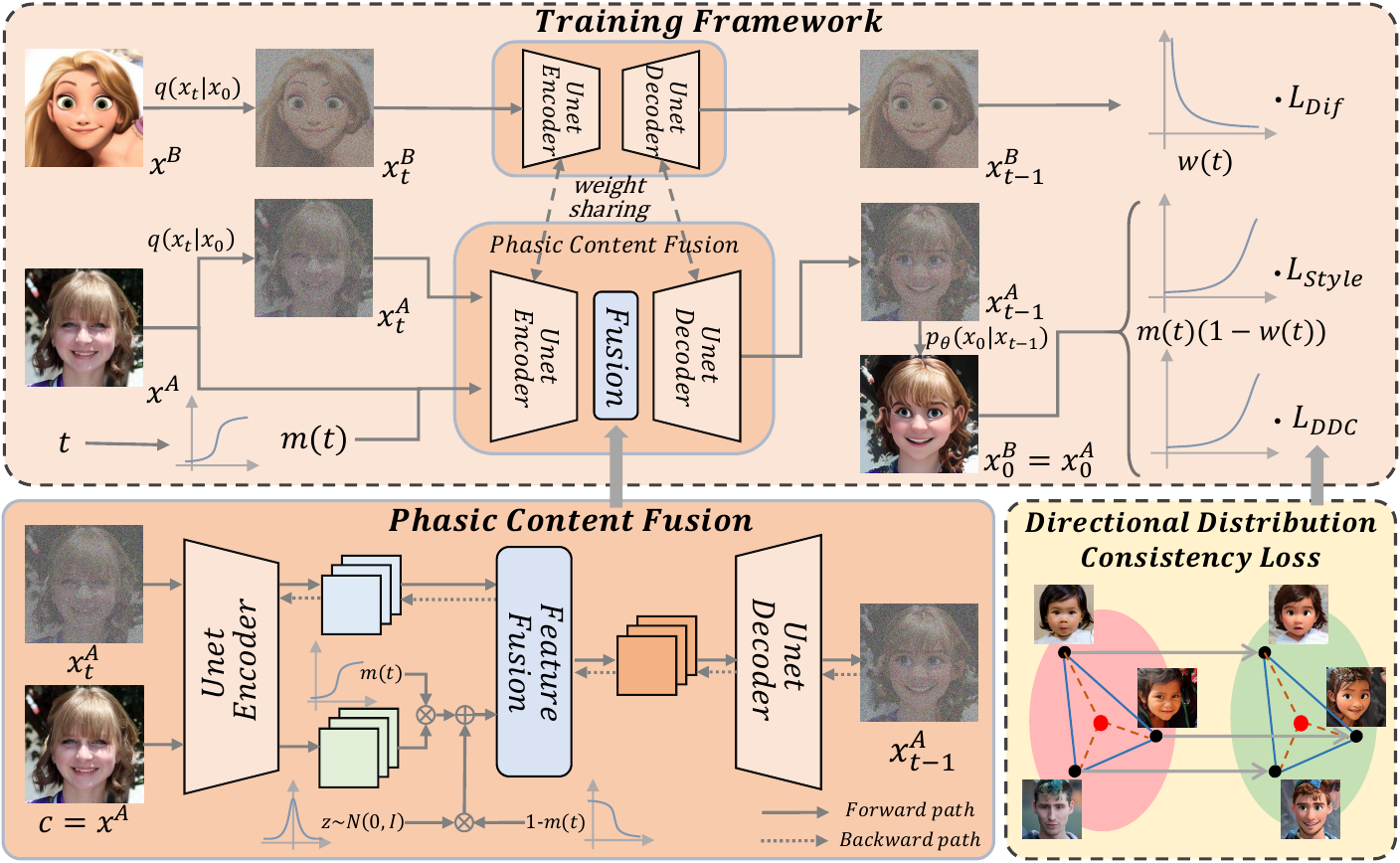}
\caption{\hut{Model Framework. The training of our model is explicitly decomposed into two stages: learn content information and style transfer at $t$-large stage (beginning denoising steps), and learn local details in the target domain at $t$-small stage. We design two training paths, the shifted sigmoid function $m(t)$ and a weighting function $w(t)$ to facilitate the phasic training. 
With the help of our phasic content fusion module and directional distribution consistency loss, our model can keep content well and avoid overfitting problem.}}
\label{fig:model framework}
\vspace{-0.1in}
\end{figure*}
\textbf{Diffusion Model.}
Denoising diffusion probabilistic models (DDPM)~\cite{ho2020denoising} has acheived high quality image generation without adversarial training~\cite{zhang2020apb2face,zhang2020freenet}.The key point of diffusion model is that assume forward process as Markov process that gradually adds noise to input image and use neural network to predict added noise to complete backward process and image reconstruction.

Given a source data distribution $x_{0} \sim q\left(x_{0}\right),\beta_{t} \in(0,1)$, diffusion model defines the forward process by:
\begin{equation}
   \begin{aligned}
        &q\left(x_{1}, \ldots, x_{T} \mid x_{0}\right)  :=\prod_{t=1}^{T} q\left(x_{t} \mid x_{t-1}\right), \\
        &q\left(x_{t} \mid x_{t-1}\right):=\mathcal{N}\left(x_{t} ; \sqrt{1-\beta_{t}} x_{t-1}, \beta_{t} \mathbf{I}\right).
    \end{aligned} 
\end{equation}

 And the backward process is approximated through a neural network to generate an image from the Gaussian noise $X_T \sim \mathcal{N}(0,I)$ iteratively by:
\begin{equation}
p_{\theta}\left(x_{t-1} \mid x_{t}\right):=\mathcal{N}\left(x_{t-1} ; \mu_{\theta}\left(x_{t}, t\right), \Sigma_{\theta}\left(x_{t}, t\right)\right),
\label{denoising}
\end{equation}
where $\mu_{\theta}(x_t,t)$ and $\Sigma_{\theta}\left(x_{t}, t\right)$ (setted as a constant in DDPM~\cite{ho2020denoising}) are predicted by the neural network. 

To futher improve the diffusion model, recent works have made great progress in  accelerating denoising process~\cite{song2020denoising} and improving generation quality~\cite{nichol2021improved,dhariwal2021diffusion}.
With flexible controlling ability of sampling process in diffusion model, it has also been employed in different sub-tasks of image generation like image-to-image translation and text-to-image generation, achieving an overwhelming performance~\cite{9878402,9746901,kwon2022diffusion,su2022dual,zhao2022egsde}. These applications show great potential of diffusion model in conditional image generation, but they all face the overfitting problem when the training samples are limited. And there is still a lack of diffusion models which focusing on scenarios with few-shot training samples. Thus, we propose a novel few-shot diffusion model with phasic content fusion and directional distribution consistency loss which can avoid overfitting problem and keep content information well.


\textbf{Few-shot Image Generation.}
The goal of few-shot image generation is to produce high-quality and diverse images in a new domain with only a small number of training data. Directly fine-tuning a pre-trained GAN is a common and straightforward approach~\cite{bartunov2018few, clouatre2019figr, liang2020dawson, wang2018transferring}. However, this often leads to model overfitting if the entire network is fine-tuned. Researchers have found that modifying only part of the network weights~\cite{mo2020freeze, wang2018transferring} and using different types of regularization~\cite{li2020few, zhang2019consistency}, along with batch statistics~\cite{noguchi2019image} can prevent overfitting. Data augmentation techniques have also been utilized to increase the amount of training data and enhance the robustness of the generative model~\cite{tran2021data, zhao2020differentiable, zhao2020image}. But it's still hard for these models to train on a dataset with less than 10 samples. Recently, IDC~\cite{ojha2021few} and RSSA~\cite{xiao2022few} introduced two new loss functions to keep the structure of the generated distribution. However, there is a lack of analysis on the proposed loss functions, which can be further improved and they also face the problem of content missing due to the lack of content maintenance. To solve these problems, we take a deep insight into loss functions in IDC and RSSA and propose a novel directional distribution consistency loss, which improves the training stability and effectiveness. Moreover, with our phasic content fusing module and iterative cross-domain structure guidance strategy, our model can keep the structure information well during domain adaptation compared to the existing methods.
\section{Method}

We propose a novel few-shot diffusion model with phasic content fusion and effective directional distribution consistency loss.
\yi{Given a diffusion model $\epsilon^A_{\theta}(x_t, t)$ pretrained on source domain $A$, we train a few-shot diffusion model $\epsilon_{\theta}(x_t, t)$ on target domain $B$, using $\epsilon^A_{\theta}(x_t, t)$ as initialization.}
\yi{During inference stage,} our model takes an image $x^A$ from source domain $A$ as input, 
we first sample the start point $x_{t}^A$ through the forward process $q(x_t|x_0)$ (\yi{adds Gaussian noise}). Then, with our few-shot diffusion model $\epsilon_{\theta}(x_t, t)$ \yi{trained on target domain}, we iteratively predict $x^A_{t-1}$ from $x^A_{t}$ by the denoising process $p_{\theta}(x_{t-1}|x_t)$ to get the final output $x^{A\to B}=x_0^{A}$, \yi{which is transferred to the target} domain and \yi{keeps} original content information of $x^A$.

To better \yi{learn} the content in source domain and local details in target domain, we explicitly decompose the training process into two stages: \yi{the first stage learns} content and style information at $t$-large and \yi{the second stage learns} target-domain local details at $t$-small. 
Additionally, we introduce a \yi{phasic} content fusion module, which adaptively incorporates content information into our model based on the current learning stage \yi{($t$)}, resulting in improved capture of content information. 
Moreover, to solve the overfitting problem, we propose \yi{a novel} directional distribution consistency loss, which 
\yi{uses directional guidance to enforce the structure of the generated distribution to be similar to source distribution, while the center close to that of the target distribution, and effectively avoids distribution rotation during training}. 
Lastly, by employing our iterative cross-domain structure guidance strategy \yi{during inference stage}, our model effectively maintains \yi{the} structure in source image, enhancing consistency of generated and \yi{input} images in terms of structure and outline.

\subsection{Training with Phasic Content Fusion}

\begin{figure}[t]
\centering
\includegraphics[width=0.45\textwidth]{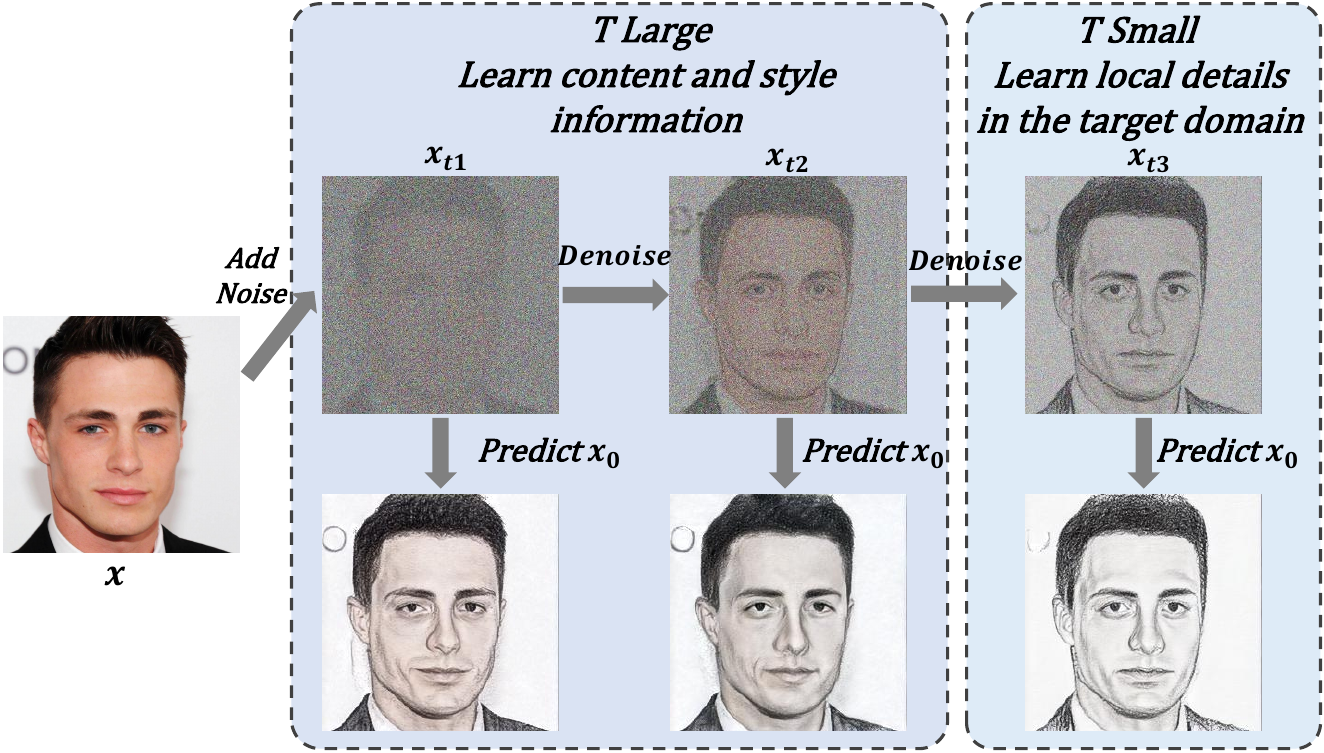}
\caption{Our phasic training strategy learns the content and style information at t-large, while learns local details in the target domain (sketch here) at t-small.}
\label{fig:phasic trainig strategy}
\vspace{-0.1in}
\end{figure}
\textbf{Phasic Training Strategy.}
\hut{Diffusion model learns different information in different training stages according to time step t~\cite{choi2022perception}, i.e., learn contents at t-large while learn details at t-small. When t is small, it's hard to change both the content and style. Therefore, directly training diffusion model with the loss function in few-shot GAN~\cite{ojha2021few,xiao2022few} leads to failure in style transfer at t-small, causing inaccurate capture of style\cite{zhu2022few} as Fig.~\ref{fig:figure 1} shows.}

To solve this problem, \hut{we expect} our diffusion model \hut{to capture} the content and style information at t-large, while only learn the local details of target domain at t-small (as Fig.~\ref{fig:phasic trainig strategy} shows). 
\hutnew{We decompose the training into two stages, i.e., t-large stage to learn content and style, and t-small stage to learn local details of target domain. To accomplish this goal, we first design a two-path training framework: apart from the training path on target domain, we introduce another training path that incorporates source domain images to provide content guidance and better learn the content at t-large. Then we} introduce a shifted sigmoid function $m(t)=\frac{1}{1+e^{-(t-T_s)}}$ \hutnew{and a weighting function $w(t)=1-(\frac{t}{T})^{\alpha}$, and integrate them into the model structure and loss functions to enforce larger weight to content and style related learning at t-large, and larger weight to target domain local details learning at t-small.}

\textbf{Phasic Content Fusion Module.} 
\yi{For the training path that incorporates source domain images to better learn content at $t$-large, the inputs contain both noised image $x_t^A$ and source image $x^A$, where the latter is used to supplement the missing content in $x_t^A$ when $t$ is large.}
\hutnew{
We propose a novel content fusion module 
to adaptively fuse the content of $x^A$ into our model with $m(t)$ as weight, i.e., more content is fused when $t$ is large.}

\hutnew{Specifically,} the phasic content fusion module is based on the UNet in diffusion model. 
We employ the UNet encoder to extract image features \hutnew{$E(x^A)$} and $E(x_t^A)$. 
\yi{Since content is learnt more in the beginning denoising steps ($t$-large), the influence of content in $x_A$ should be increased when $t$ is large and lowered when $t$ is small.
We accomplish this goal by adaptively fusing the content feature $E(x^A)$ and noise $z \sim\mathcal{N}(0,I)$ using $m(t)$ as the weight for content, i.e.,}
\hutnew{
 $\hat{E}(x^A)=m(t)E(x^A)+(1-m(t))z$.} 
 \hut{Then, we \yi{further} fuse $\hat{E}(x^A)$ with $E(x^A_t)$ using several convolution blocks to get the fused feature $\hat{E}(x^A,x^A_t)$. At last, we feed the fused feature to UNet decoder to predict the noise $\epsilon_t$ and obtain $x^A_{t-1}$, which contains the enhanced content information.}

\subsection{Directional Distribution Consistency }
\begin{figure}[t]
\centering
\includegraphics[width=0.45\textwidth]{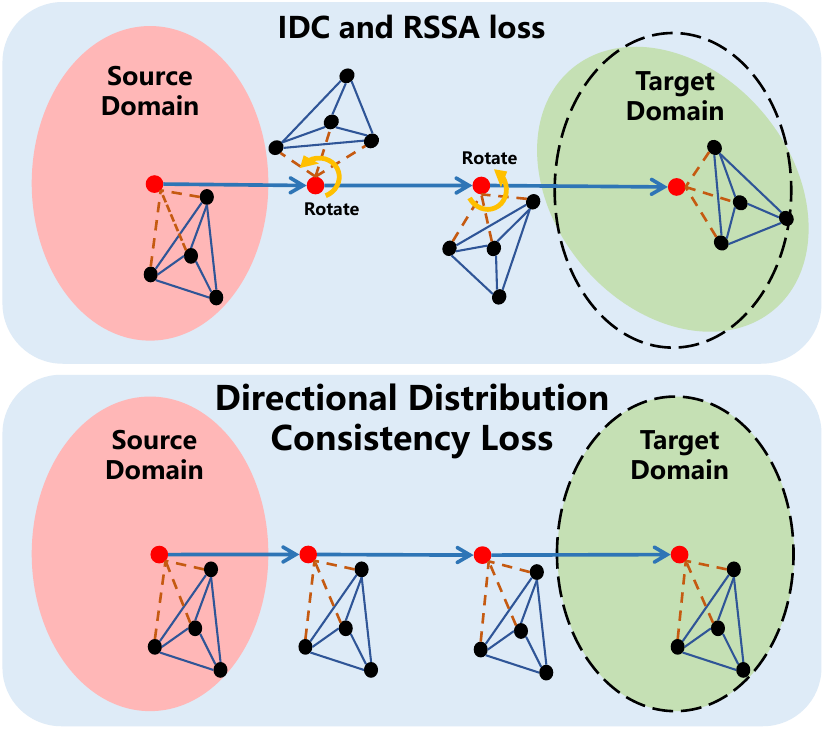}
\caption{Compare our DDC loss with IDC and RSSA: our DDC loss explicitly constrain the structure of generated distribution while IDC and RSSA may suffer from distribution rotation in training process, \hut{which interferes training stability and efficiency.}}
\label{fig:cddc loss}
\vspace{-0.1in}
\end{figure}

\hut{In this section, we introduce our training losses to keep structure of generated distribution and transfer the style.}

\textbf{Directional distribution consistency loss.} 
In the few-shot scenario, model is highly susceptible to overfitting. \yi{To cope with overfitting}, IDC~\cite{ojha2021few} and RSSA~\cite{xiao2022few} propose new loss functions to maintain the structure of generated distribution by constraining the similarity between source and generated distributions in a training batch. We theoretically prove that the final goal of their loss functions is to keep the structure and \yi{scale} of the generated distribution the same as the source distribution, while sharing the same center with target \yi{distribution} (refer to Appendix). 
\hut{However, although they can avoid the generation drift problem, \yi{they only require the pairwise distances of generated samples in target and source domains to be similar, which leads to} distribution rotation during the training process as Fig.~\ref{fig:cddc loss} shows, \yi{and may cause} unstable and ineffective training.}

\begin{figure}[t]
\centering
\includegraphics[width=0.45\textwidth]{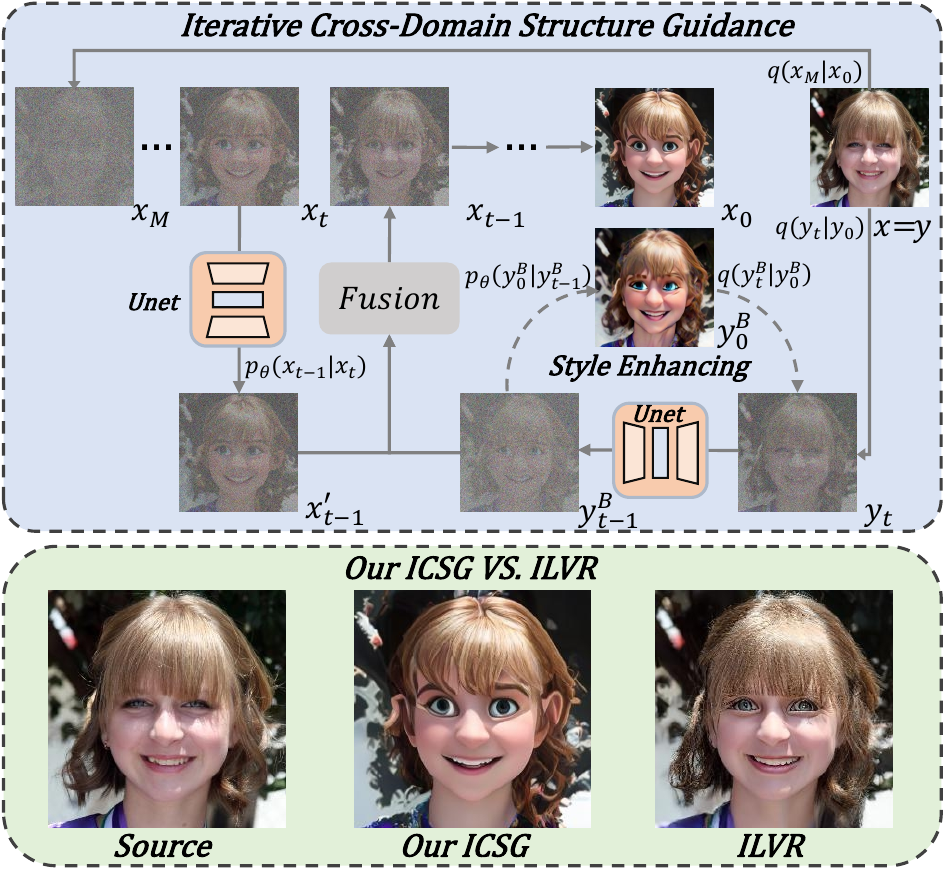}
\caption{\hut{Process of our iterative cross-domain structure guidance strategy (ICSG) and comparison with ILVR\cite{choi2021ilvr}, where ILVR tends to reconstruct the source image and lose the style information.}}
\label{fig:ICSG}
\vspace{-0.1in}
\end{figure}

\hut{To avoid \yi{distribution rotation during training}, we propose a new directional distribution consistency loss (DDC). Compared to the existing loss functions, our DDC loss \yi{introduces a directional guidance to} directly optimizes the final goal (\yi{distribution} structure maintenance and center \yi{movement}), which avoids the generated distribution from rotation and improves the training efficiency.}


 Specifically, \yi{given} the source dataset $A=\{x^A_1,\cdots,x^A_n\}$ and target dataset $B=\{x^B_1,\cdots x^B_m\}$, we extract the image features by image encoder $E$ \yi{for each dataset}. Then we compute the cross-domain direction vector $w$ from the center of \yi{source domain} to the center of \yi{target domain in feature space} by:
\begin{equation}
   \begin{aligned}
        w=\frac{1}{m}\sum_{i=1}^{m} E(x^B_i)-\frac{1}{n}\sum_{i=1}^{n} E(x^A_i).
    \end{aligned} 
    \label{eq:cross-domain vector}
\end{equation}

\yi{We} leverage the \yi{directional} vector $w$ to constrain the structure of the generated distribution to match that of original distribution, while also ensure its center coincides with that of the target distribution, \yi{by the following directional distribution consistency loss:} 
\begin{equation}
   \begin{aligned}
        \mathcal{L}_{DDC}=\|E(x^A)+w,E(x_0^{A\to B})\|^2,
    \end{aligned} 
    \label{eq:DDC loss}
\end{equation}
\hut{where $x^A$ is the source image and $x_0^{A\to B}$ is the output image in target domain.}
Through this \yi{loss}, we explicitly enforce consistency of the spatial structure between the generated and original distributions during domain adaptation (as Fig.~\ref{fig:cddc loss} shows).

We employ CLIP \yi{as the encoder $E$} to embed the images, since CLIP has been proved to be an effective encoder to extract features from different domains~\cite{vinker2022clipasso},  
\yi{which} can help distinguish between the domain-specific and domain-independant features.


\textbf{Style loss.} To better capture the style information, we \yi{adopt a} style loss \yi{which averages the Gram matrix~\cite{gatys2016image} based style difference} between our generated image \hut{$x^{A\to B}_0$} and target images $B=\{x^B_{1},\cdots x^B_{m}\}$ by:
\begin{equation}
   \begin{aligned}
        \mathcal{L}_{style}=\frac{1}{m}\sum_{i=1}^{m}\sum\limits_l w_l\|G^l(x_0^{A\to B})-G^l(x_i^B)\|^2,
    \end{aligned} 
    \label{eq:cross-domain vector}
\end{equation}
where $G^l$ is the Gram matrix and $m\le 10$.

\textbf{Diffusion Loss.} At last, we inherit the loss function in DDPM~\cite{ho2020denoising} to help train our diffusion model on the target domain $B$ \hut{without the content fusion module}:
\begin{equation}
    \begin{aligned}
    \mathcal{L}_{dif}=||\epsilon_{\theta}(x^B_t,t)-\epsilon||^2.
    \label{eq:diffusion loss}
    \end{aligned}
\end{equation}

\textbf{Total loss.} With the above three loss functions, the final loss function $\mathcal{L} $ is calculated by:
\begin{equation}
    \begin{aligned}
    \mathcal{L}=
    &m(t) (1-w(t)) (\lambda_{DDC}\mathcal{L}_{DDC}(x^A,x^{A\to B}_0)+\\
    &\lambda_{style}\mathcal{L}_{style}(x^{A\to B}_0,x^B))+
    w(t)\mathcal{L}_{dif}(x^B),
    \label{eq:total loss}
    \end{aligned}
\end{equation}
where $\lambda$s are the hyperparameters, $m(t)$ is the shifted sigmoid function and $w(t)$ is the weight balancing function.
\begin{figure*}[t]
\centering
\includegraphics[width=0.88\textwidth]{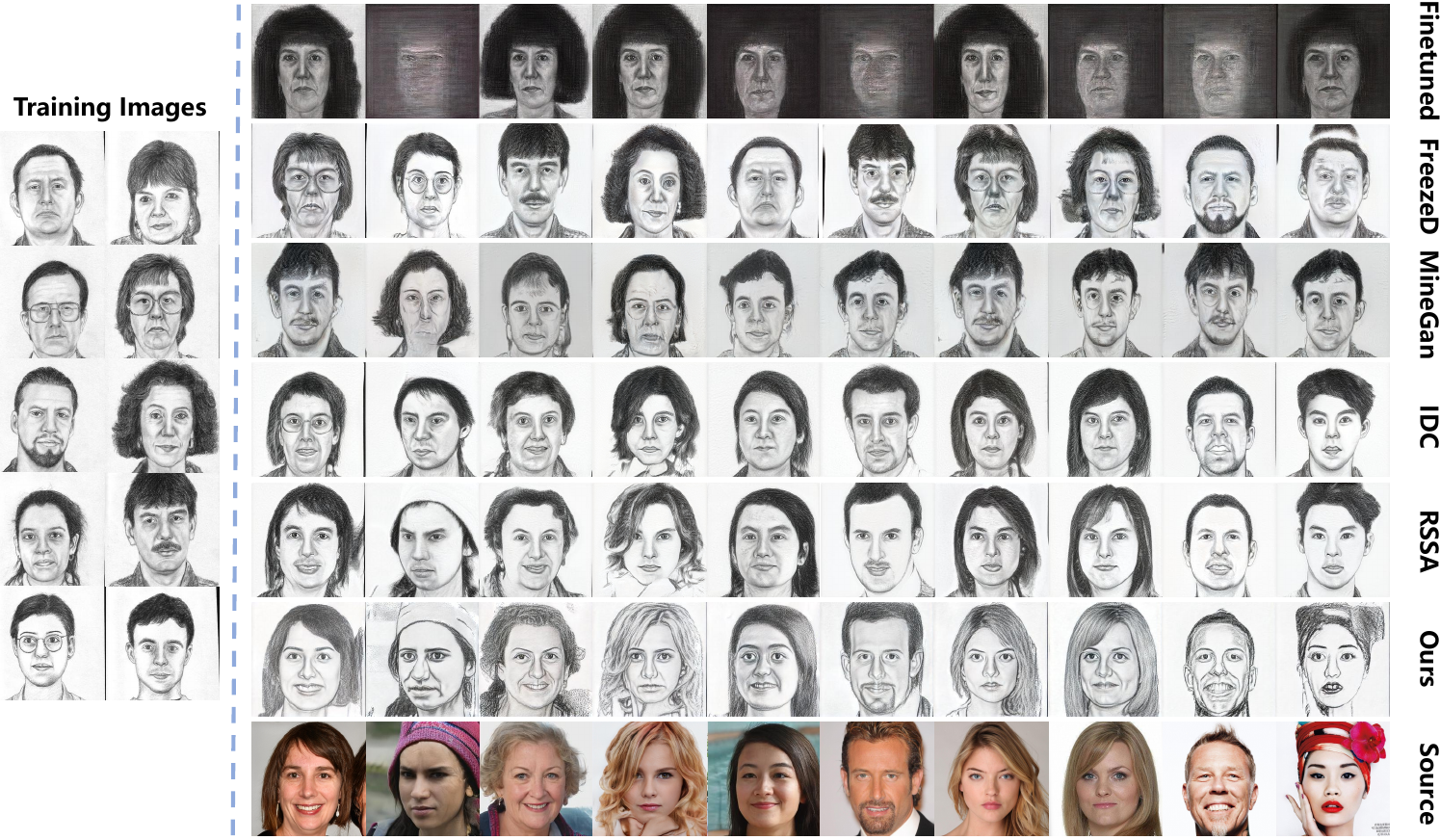}
\caption{Comparison results on sketches: our model achieves good performance in both content maintenance and domain adaptation.}
\label{fig:Comparison results on sketches}
\vspace{-0.1in}
\end{figure*}

\begin{figure*}[t]
\centering
\includegraphics[width=0.88\textwidth]{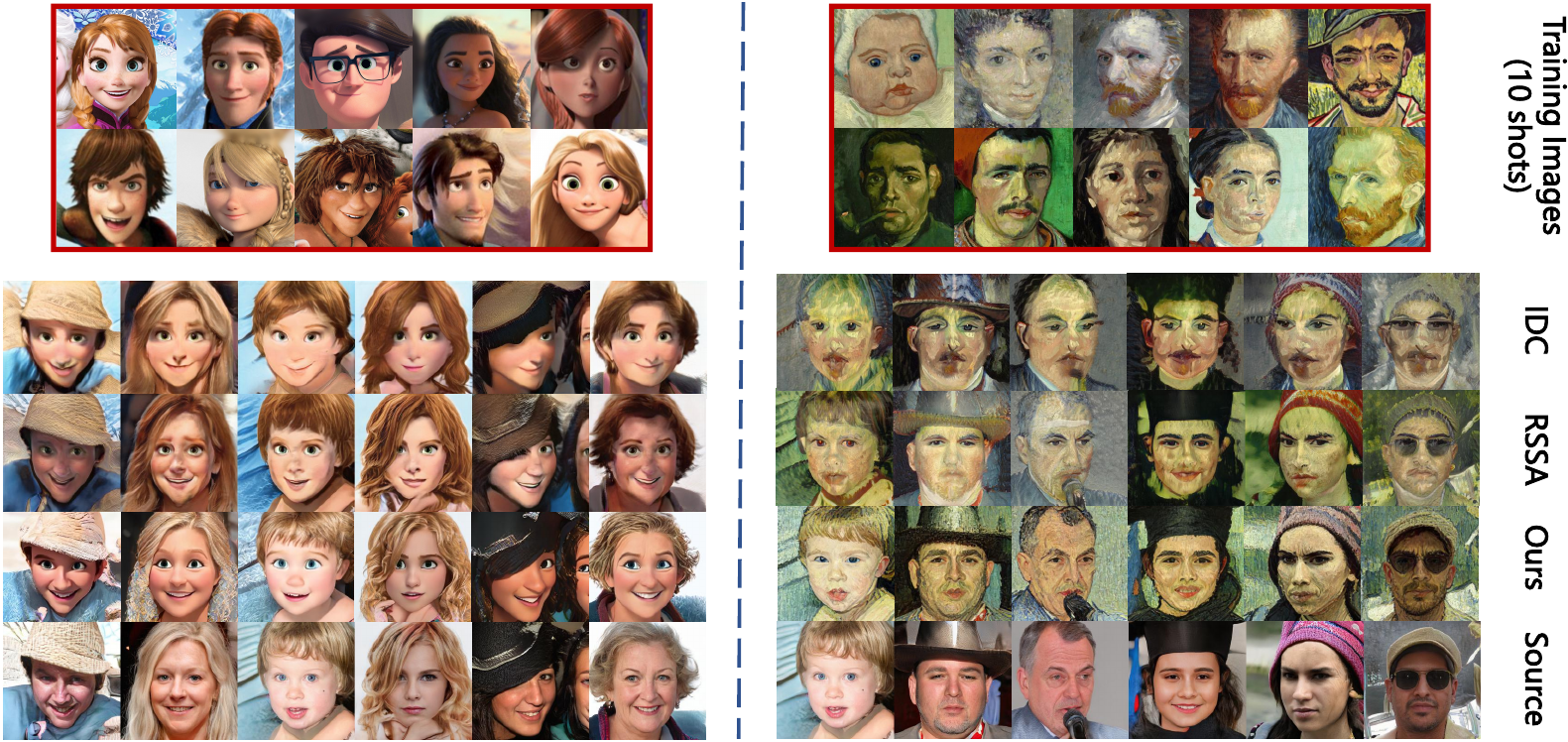}
\caption{Comparison results on Cartoon and Van Gogh painting dataset with IDC and RSSA.}
\label{fig:Comparison results on cartoon and vangoah}
\vspace{-0.1in}
\end{figure*}
\subsection{Iterative Cross-domain Structure Guidance}
Our proposed phasic content fusion module in the network can help keep the content information well. But there is still a room to improve \hut{the \yi{preservation of} local structures \yi{in} the source image \yi{during} the inference stage}.
We propose a novel iterative cross-domain structure guidance strategy (ICSG), which constantly enhances the local structures and keeps the style unchanged during the denoising process.



ILVR~\cite{choi2021ilvr} \yi{proposes a conditioning method to generate images with similar semantics to a reference image, where the downsampled image $\phi_N(x_0)$ of the generated image $x_0$ is pulled close to the downsampled image $\phi_N(y)$ of the reference image $y$ ($\phi_N$ is a linear low-pass filter). At each time step $t$, ILVR denoises $x_t$ to $x_{t-1}$ with a local condition where $\phi_N(x_{t-1})$ and $\phi_N(y_{t-1})$ are similar: $x_{t-1}=x'_{t-1}+\phi_N(y_{t-1})-\phi_N(x'_{t-1})$, $x'_{t-1}\sim p_{\theta}(x'_{t-1}|x_t)$
We can apply ILVR to our task by using the source image $x$ as the reference image. But since the target domain is different in style from the source domain, directly applying ILVR leads to shifted style (Fig.~\ref{fig:ICSG}).}

\yi{To address the above problem, }
\hut{
we propose our iterative cross-domain structure guidance strategy (ICSG) as Fig.~\ref{fig:ICSG} shows.} 
\yi{In our case, the reference image $y$ is a source image $x$. 
Instead of directly sampling $y_{t-1}$ via the forward process $q(y_{t-1}|y_0)$, we obtain a target domain style $y^B_{t-1}$ by first sampling $y_t \sim q(y_{t}|y_0)$ and then translating it to target domain $y^B_{t-1}$ by using our trained diffusion model $p_{\theta}(y_{t-1}|y_t)$. We then enforce structure similarity between $\phi_N(x_{t-1})$ and $\phi_N(y^B_{t-1})$ by: }
\begin{equation}
\begin{aligned}
\yi{x_{t-1} = x'_{t-1}+\phi_N(y^B_{t-1})-\phi_N(x'_{t-1}), x'_{t-1}\sim p_{\theta}(x'_{t-1}|x_t).}
\end{aligned}
\end{equation}
\yi{Compared to ILVR, our ICSG can eliminate the interference from source style and better preserve the structure.}

\yi{We further enhance the target domain style of $y^B_{t-1}$ by iteratively applying a Style Enhancement (SE) module, which repeats}
\hut{
the following steps: 
\yi{(1) compute $y^B_0$ from $y^B_{t-1}$ by $p_{\theta}(y^B_0|y^B_{t-1})$ with $\epsilon_{\theta}(y^B_t, t)$ in last $p_{\theta}(y^B_{t-1}|y^B_t)$,}
(2) add $t$-step noise into \yi{$y^B_0$ to get new $y^B_t \sim q(y^B_t|y^B_0)$,}
and (3) denoise \yi{$y^B_t$ to $y^B_{t-1}$ by our model $p_{\theta}(y^B_{t-1}|y^B_t)$. }
\yi{We apply the Style Enhancement (SE) module for $M$ times ($M$ depends on the style gap between source and target domain) until $y^B_{t-1}$ is fully transferred to the target domain style.}}


\begin{figure*}[t]
\centering
\includegraphics[width=0.88\textwidth]{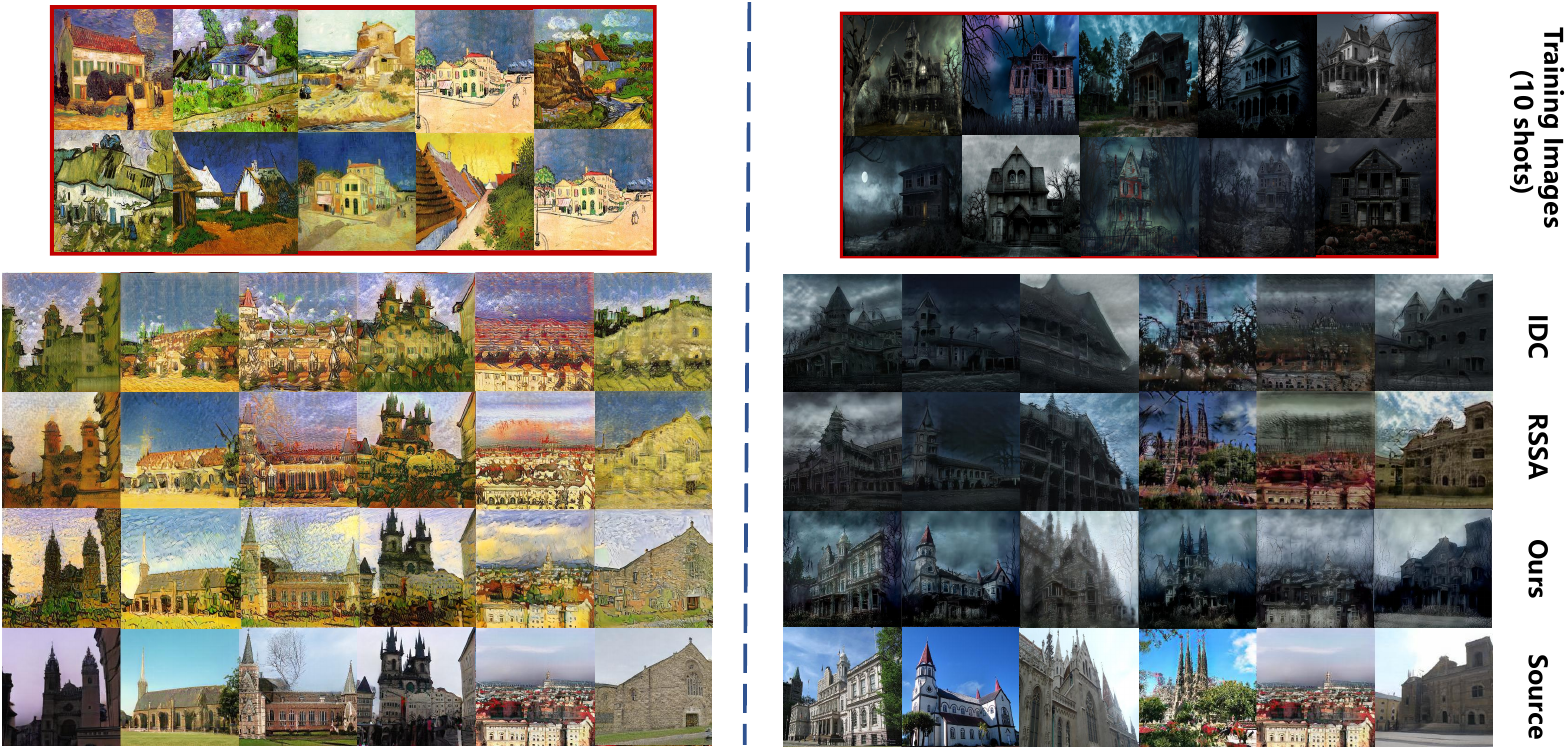}
\caption{Comparison results on haunted houses and village painting by Van Gogh with IDC and RSSA.}
\label{fig:Comparison results on haunted houses and village painting by Van Gogh}
\vspace{-0.1in}
\end{figure*}

\begin{table*}[t]

\centering
\footnotesize
\scalebox{0.9}{
\begin{tabular}{llllllllllll}
\hline
\multirow{2}{*}{Metric}           & \multirow{2}{*}{Method} & \multicolumn{2}{c}{FFHQ $\to$ Sketches} & \multicolumn{2}{c}{FFHQ $\to$ Cartoon} & \multicolumn{2}{c}{FFHQ $\to$ Van. face} & \multicolumn{2}{c}{Church $\to$ Van. vil} & \multicolumn{2}{c}{Church $\to$ Haunted} \\ \cline{3-12} 
                                  &                         & 10-shot            & 5-shot             & 10-shot            & 5-shot            & 10-shot             & 5-shot             & 10-shot             & 5-shot              & 10-shot             & 5-shot             \\ \hline
\multirow{6}{*}{IS $\uparrow$}    & FreezeD                 & 1.502              & 1.636              & 3.047              & 2.205             & 1.333               & 1.784              & 1.795               & 2.331               & 2.527               & 1.949              \\
                                  & MineGan                 & 1.320              & 1.700              & 2.343              & 2.917             & 1.604               & 1.710              & 2.412               & 2.080               & 2.241               & 2.282              \\
                                  & IDC                     & 1.640              & 2.100              & 2.829              & 2.100             & 1.373               & 1.736              & 2.798               & 2.945               & 2.768               & 2.434              \\
                                  & RSSA                    & 1.875              & 2.135              & \textbf{3.595}     & 3.098             & 2.129               & 1.983              & \textbf{3.139}      & 3.058            & 2.634               & 2.598              \\
                                  & fine-tune               & 1.871              & 1.532              & 1.838              & 1.725             & 1.957               & 1.901              & 2.856               & 2.724               & 1.618               & 1.324              \\
                                  & Ours                    & \textbf{2.361}     & \textbf{2.146}     & 3.410              & \textbf{3.317}    & \textbf{2.449}      & \textbf{2.134}     & 3.072               & \textbf{3.088}               & \textbf{2.784}      & \textbf{2.657}     \\ \hline
\multirow{6}{*}{IC-LPIPS $\uparrow$} & FreezeD                 & 0.351              & 0.345              & 0.472              & 0.467             & 0.506               & 0.462              & 0.328               & 0.343               & 0.485               & 0.405              \\
                                  & MineGan                 & 0.340              & 0.319              & 0.431              & 0.526             & 0.468               & 0.452              & 0.559               & 0.368               & 0.486               & 0.497              \\
                                  & IDC                     & 0.418              & 0.542              & 0.575              & 0.557             & 0.574               & 0.524              & 0.666               & 0.655               & 0.623               & 0.602              \\
                                  & RSSA                    & 0.478              & 0.471              & 0.590              & 0.582             & 0.619               & 0.598              & \textbf{0.679}      & 0.671               & 0.623               & 0.625              \\
                                  & fine-tune               & 0.469              & 0.332              & 0.362              & 0.337             & 0.411               & 0.373              & 0.414               & 0.195               & 0.161               & 0.258              \\
                                  & Ours                    & \textbf{0.557}     & \textbf{0.551}     & \textbf{0.630}     & \textbf{0.637}    & \textbf{0.625}      & \textbf{0.606}     & 0.655               & \textbf{0.673}               & \textbf{0.666}      & \textbf{0.691}     \\ \hline
\multirow{6}{*}{SCS $\uparrow$}   & FreezeD                 & 0.288              & 0.291              & 0.376              & 0.350             & 0.366               & 0.369              & 0.356               & 0.356               & 0.196               & 0.234              \\
                                  & MineGan                 & 0.289              & 0.296              & 0.386              & 0.400             & 0.373               & 0.426              & 0.397               & 0.394               & 0.287               & 0.294              \\
                                  & IDC                     & 0.338              & 0.475              & 0.516              & 0.475             & 0.560               & 0.496              & 0.557               & 0.484               & 0.458               & 0.297              \\
                                  & RSSA                    & 0.496              & 0.504              & 0.715              & 0.707             & 0.702               & 0.631              & 0.715               & 0.695               & 0.649               & 0.637              \\
                                  & fine-tune               & 0.179              & 0.293              & 0.246              & 0.353             & 0.335               & 0.342              & 0.259               & 0.313               & 0.268               & 0.286              \\
                                  & Ours                    & \textbf{0.623}     & \textbf{0.653}     & \textbf{0.837}     & \textbf{0.842}    & \textbf{0.811}      & \textbf{0.802}     & \textbf{0.838}      & \textbf{0.826}      & \textbf{0.840}      & \textbf{0.829}     \\ \hline
\end{tabular}
}
\vspace{0.05in}
\caption{Quantitative comparison on IS, IC-LPIPS and SCS with differnet source and target domains. Our model outperforms the existing methods in both generating quality (higher IS) and diversity (higher IC-LPIPS and SCS).}
\vspace{-0.1in}
\label{tab:quantitative comparison}
\end{table*}
\section{Experiments}


\subsection{Experiment Settings}
We compare our model with the existing few-shot generation models: FreezeD~\cite{mo2020freeze} , MineGAN~\cite{wang2020minegan} , IDC~\cite{ojha2021few} and RSSA~\cite{xiao2022few}, where IDC and RSSA are the state-of-the-art method. For a fair comparison, we employ StyleGAN2~\cite{karras2020analyzing} as the backbone for all these methods. Moreover, to validate the effectiveness of our method, we fine-tune a diffusion model which shares the same settings as ours.

We conduct experiments on two datasets: (1) Flickr-Faces-HQ (FFHQ)~\cite{karras2019style} and (2) LSUN Church~\cite{yu2015lsun}. And we translate the model to the target domain: (1) Sketches~\cite{wang2008face}, (2) Cartoon~\cite{pinkney2020resolution}, (3) Paintings by Van Gogh~\cite{yaniv2019face} and (4) Haunted houses~\cite{ojha2021few}. The experiments are conducted in both 10-shot and 5-shot settings.

\textbf{Evaluation protocals.} We employ \yi{three} metrics to evaluate model performance:
    (1) \textbf{IS:} Inception Score~\cite{barratt2018note} measures the high resolution and diversity of images by calculating the information entropy of the generated images.
    (2) \textbf{IC-LPIPS:} Intra-cluster pairwise LPIPS distance~\cite{ojha2021few} first classifies generated images into $k$ clusters according to their LPIPS distance to the $k$ target samples. By averaging the mean LPIPS distance to the corresponding target samples in each cluster, a higher IC-LPIPS indicates a better generation diversity.
    (3) \textbf{SCS:} Structural Consistency Score~\cite{xiao2022few} first extracts edge maps of pairwise source and generated images by HED~\cite{xie2015holistically} and then measures the mean similarity score between them. Higher SCS indicates better spatial structural consistency between source and generated distribution, leading to higher diversity of generated images.

\subsection{Performance Evaluation}
\textbf{Qualitative Evaluation.}
We first compare the visual quality of the generated images on sketch domain. We randomly sample 5 source images from the offered latent code in IDC~\cite{ojha2021few} and 5 images from CelebA-HQ~\cite{karras2018progressive}. 
Fig.~\ref{fig:Comparison results on sketches} shows the comparison results. It can be seen that FreezeD, MineGAN and the fine-tuned diffusion model are all overfitted whose results have poor relation to the source images. Both IDC and RSSA can keep part of features in the source images, but there are still some content missing, especially when dealing with CelebA-HQ images. Compared to them, our method keeps the content well while translating images to the target domain.

To further validate the effectiveness of our model, we compare our model with the state-of-the-art method: IDC and RSSA on more datasets. Besides sketches, we conduct experiments on cartoon and Van Gogh painting with the pretrained model on FFHQ in Fig.~\ref{fig:Comparison results on cartoon and vangoah}. And we also compare the performance when translating from LSUN church to haunted houses and village painting by Van Gogh in Fig.~\ref{fig:Comparison results on haunted houses and village painting by Van Gogh}. All the results show that our model can maintain the content information and translate the domain well.

\textbf{Quantitative Evaluation.}
We quantitatively compare our model with the state-of-the-art methods on 5 domain adaption experiments: FFHQ to sketches, FFHQ to Cartoon, FFHQ to Van Gogh painting, LSUN Church to Van Gogh painting and LSUN Church to hunted house. We conduct the experiments on both 5-shot and 10-shot settings. Specifically, We first sample 1000 images from StyleGAN2~\cite{karras2020analyzing} as the source images and generate 1000 images in target domain by all the methods. Then we calculate the IS, IC-LPIPS and SCS on these generated images in Tab.~\ref{tab:quantitative comparison}. For the content keeping metrics IC-LPIPS, SCS and the generation quality metric IS, our model outperforms the existing methods in almost all experiment settings. 

\begin{figure}[t]
\centering
\vspace{-0.2in}
\includegraphics[width=0.45\textwidth]{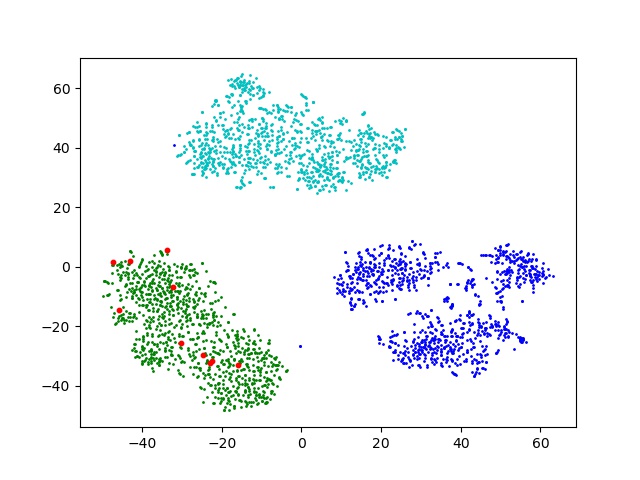}
\vspace{-0.1in}
\caption{\yi{t}-SNE results of few-shot samples (red); source images (blue); our generated results (green) and IDC generated results (cyan). It's clearly seen that our generated results are in the target domain and keeps the distribution structure well.}
\label{fig:t-sne}
\vspace{-0.15in}
\end{figure}

\subsection{Analysis on the DDC Loss}
In this section, we give a further insight in our DDC loss. We randomly sampled 1000 images from StyleGAN2 and translate them to the cartoon domain with our method and IDC~\cite{ojha2021few}. To validate that our generated distribution is more similar to source distribution, we employ t-SNE to visualize the distributions of the source images (blue), target 10-shot cartoon images (red), our generated images (green) and IDC generated images (cyan) in Fig.~\ref{fig:t-sne}. It can be seen that our generated distribution translates the domain well since the target images are all located in it and they share a close distribution center. The visualization result validates that our DDC loss can help the few-shot generative model to translate the distribution center and maintain the structure well.


\begin{table}[t]
\centering
\setlength{\abovecaptionskip}{4pt}
\setlength{\belowcaptionskip}{-0.2cm}
\setlength\tabcolsep{3pt}
\renewcommand{\arraystretch}{1.0}
\setlength\tabcolsep{5.0pt}
\resizebox{1.0\linewidth}{!}{
\begin{tabular}{ccc|ccc}
\toprule
\multicolumn{3}{c}{Method}&\multicolumn{3}{c}{Metric}\\
\hline
PCF&DDC &ICSG& IS $\uparrow$ &IC-LPIPS $\uparrow$& SCS $\uparrow$\\
\hline
\checkmark&&&1.886&0.581&0.625\\
&\checkmark&&2.018&0.586&0.629\\
\checkmark&\checkmark&& 2.699 &0.606 &0.690\\
\checkmark&&\checkmark&2.736&0.608&0.731\\
&\checkmark&\checkmark&2.426&0.605&0.791\\
\checkmark&\checkmark&\checkmark&\textbf{3.410}&\textbf{0.630}&\textbf{0.837}\\
\bottomrule
\end{tabular}}
\vspace{0.05in}
\caption{Ablation study on phasic content fusion module (PCF), directional distribution consistency loss (DDC) and the iterative cross-domain structure guidance strategy (ICSG) on cartoon.}
\vspace{-0.1in}
\label{tab:ablation study}
\end{table}
\subsection{Ablation Study}
\begin{figure}[t]
\centering

\includegraphics[width=0.48\textwidth]{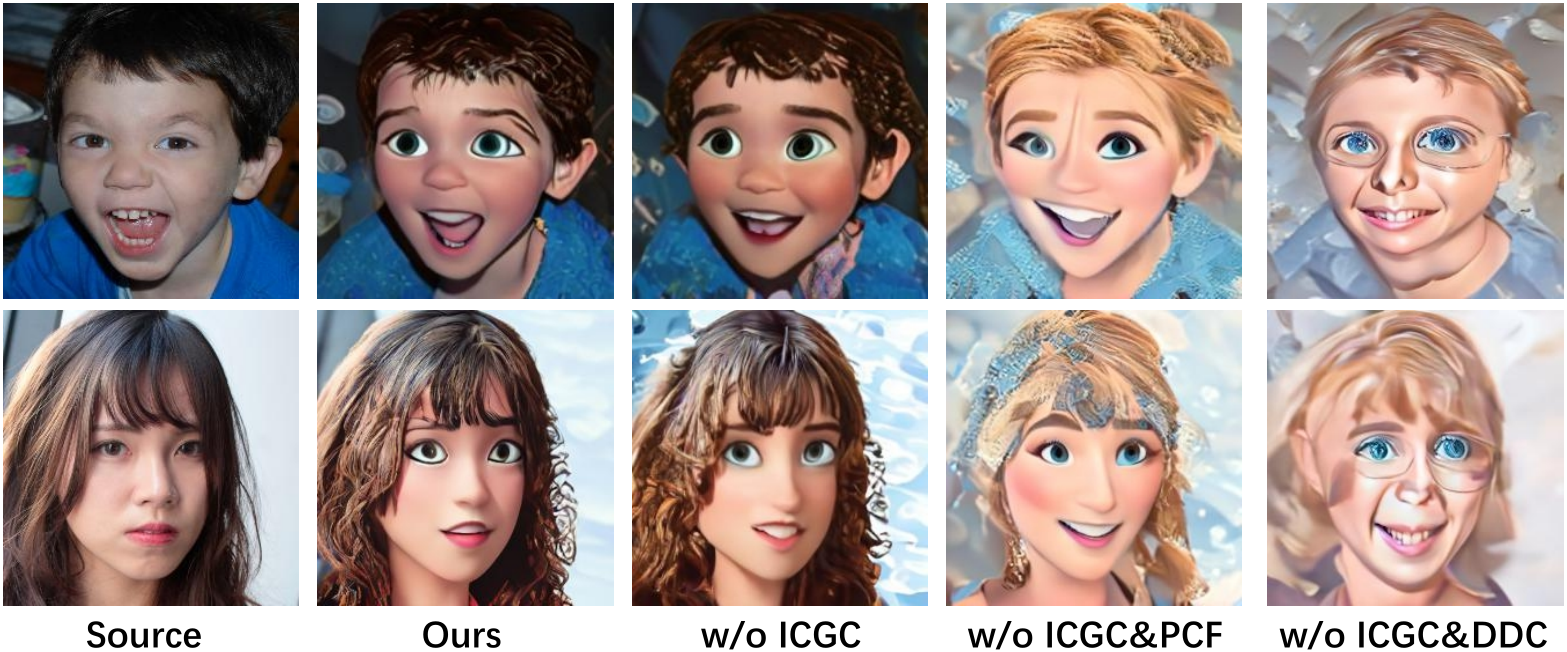}

\label{fig:ablation}
\caption{Ablation study on phasic content fusion module (PCF), directional distribution consistency loss (DDC loss) and the iterative cross-domain structure guidance strategy (ICSG) on cartoon.}
\vspace{-0.1in}
\end{figure}
To evaluate the effectiveness of our proposed methods, we conduct ablation study on the phasic content fusion module (PCF), directional distribution consistency loss (DDC loss) and the iterative cross-domain structure guidance strategy (ICSG) on the cartoon dataset. We train three networks: (1) with PCF only; (2) with DDC only and (3) with both PCF and DDC. Then, we sample 1000 images from the three models with or without ICSG respectively. We calculate 
IS, IC-LPIPS and SCS metrics for these generated images and summarize them in Tab.~\ref{tab:ablation study} and show the visualization comparison in Fig.~\ref{fig:ablation}. It can be seen that each of our proposed module is effective in either content preservation, domain translation or generation diversity.

\section{Conclusion}
In this paper, we propose a novel phasic content fusing few-shot diffusion model with directional distribution consistency loss, achieving a good performance in content preservation and few-shot domain adaption. Moreover, we propose a new iterative cross-domain structure guidance strategy which can keep the structure consistency during domain translation. Extensive quantitative and qualitative experiments show the effectiveness of our model in few-shot image generation. 

\section*{Acknowledgements}
 This work was supported by National Natural Science Foundation of China (72192821, 62272447, 61972157), Shanghai Sailing Program (22YF1420300), Shanghai Municipal Science and Technology Major Project (2021SHZDZX0102), Shanghai Science and Technology Commision (21511101200), CCF-Tencent Open Research Fund (RAGR20220121), Young Elite Scientists Sponsorship Program by CAST (2022QNRC001), Beijing Natural Science Foundation (L222117), the Fundamental Research Funds for the Central Universities (YG2023QNB17).

{\small
\bibliographystyle{ieee_fullname}
\bibliography{egbib}
}

\newpage
\appendix

\section{Overview}
\par This supplementary material consists of:
\begin{itemize}
    
\item  The implementation details on the training and testing process (Sec.~\ref{Sec:implementation details});

\item Comparison with SDEdit. (Sec. ~\ref{Sec: quantitative comparison})

\item  Ablation study on the phasic factor $T_s$ in $m(t)$, the parameters in ICSG and DDC loss (Sec.~\ref{Sec:ablation study});

\item  The theoretical analysis and more experiments on our iterative cross-domain structure guidance strategy (ICSG). (Sec.~\ref{Sec:ICSG})

\item  The theoretical analysis of our directional distribution consistency loss. (Sec.~\ref{Sec:DDC loss})
\end{itemize}
\section{Implementation Details}
\label{Sec:implementation details}
\subsection{Training Details}
In our diffusion model~\cite{ho2020denoising}, we set the maximum step to be 1000. We set the phasic factor $T_s$ in $m(t)=\frac{1}{1+e^{-(t-T_s)}}$ to 300 and the parameter $\alpha$ in $w(t)=1-(\frac{t}{T})^{\alpha}$ to 3. We start training from a pre-trained diffusion model with cosine noise schedule~\cite{nichol2021improved} on the source dataset, and fine-tune it with our phasic content fusing strategy and corresponding loss functions. Using the pre-trained Unet network, we first train our phasic content fusion model with only the diffusion loss $\mathcal{L}_{dif}$ on the source dataset, with a batch size of 8 and a learning rate of $1e^{-4}$ for 1000 iterations, to avoid interference from random weights in the early training stage.

After training the phasic content fusion module, we train the entire model with the final loss function $\mathcal{L}$ (Equation (7) in the main paper), with a batch size of 8 and a learning rate of $1e^{-4}$. We set the hyperparameters $\lambda_{DDC}$ and $\lambda_{style}$ to 1.

\subsection{Testing Details}
After training, we test our model with our iterative cross-domain structure guidance (ICSG) strategy. For the style enhancement factor $K$ in ICSG, we set $K=2$ for FFHQ~\cite{karras2019style} $\to$ Sketch~\cite{wang2008face}, and $K=1$ otherwise. Furthermore, for an input image $x$, we add 800-step noise into it as the starting point $x_M$, and employ ICSG in the denoising step until the stop step $t_{stop}$ ($t_{stop}=500$ for FFHQ~\cite{karras2019style} and  $t_{stop}=200$ for LSUN Church~\cite{yu2015lsun}). Note that a wide range of the stop step $t_steop$ and style enhancement factor $K$ have a good performance in few-shot domain adaption as illustrated in Sec.~\ref{sec:Ablation Study on ICSG}. We only choose a relatively better parameter setting in the testing stage.

\begin{table}[t]
\centering
\setlength\tabcolsep{3pt}
\renewcommand{\arraystretch}{1.0}
\setlength\tabcolsep{5.0pt}
\resizebox{1.0\linewidth}{!}{
\begin{tabular}{c|cccc|cccc}
\toprule
\multirow{2}{*}{Method}&\multicolumn{4}{c}{FFHQ $\to$ Sketches}&\multicolumn{4}{c}{FFHQ $\to$ Sketches}\\
& FID& IS $\uparrow$ &IC-L $\uparrow$& SCS $\uparrow$& FID& IS $\uparrow$ &IC-L $\uparrow$& SCS $\uparrow$\\
\midrule
SDEdit (400)&82.14&1.95&0.43&0.47&154.99&1.85&0.45&0.50 \\
SDEdit (500)&77.33&1.90&0.40&0.40&144.42&1.91&0.43&0.47 \\
SDEdit (600)&70.96&1.88&0.38&0.33&137.79&1.93&0.37&0.44 \\
PCF Only&57.62&2.11&0.52&0.51&137.79&2.70 &0.60& 0.69\\
Full Model& 47.42&\textbf{2.36}&\textbf{0.56}&\textbf{0.62}&\textbf{119.65}&\textbf{3.41}&\textbf{0.63}&\textbf{0.84}\\
\bottomrule
\end{tabular}}
\vspace{0.05in}
\caption{\textbf{Comparison results between our model and SDEdit} with different nosing steps (400, 500 and 600) on FID, IS, IC-LPIPS and SCS metrics. }
\label{tab:comparison with SDEdit}
\end{table}

\begin{figure*}[t]
\centering
\includegraphics[width=0.98\textwidth]{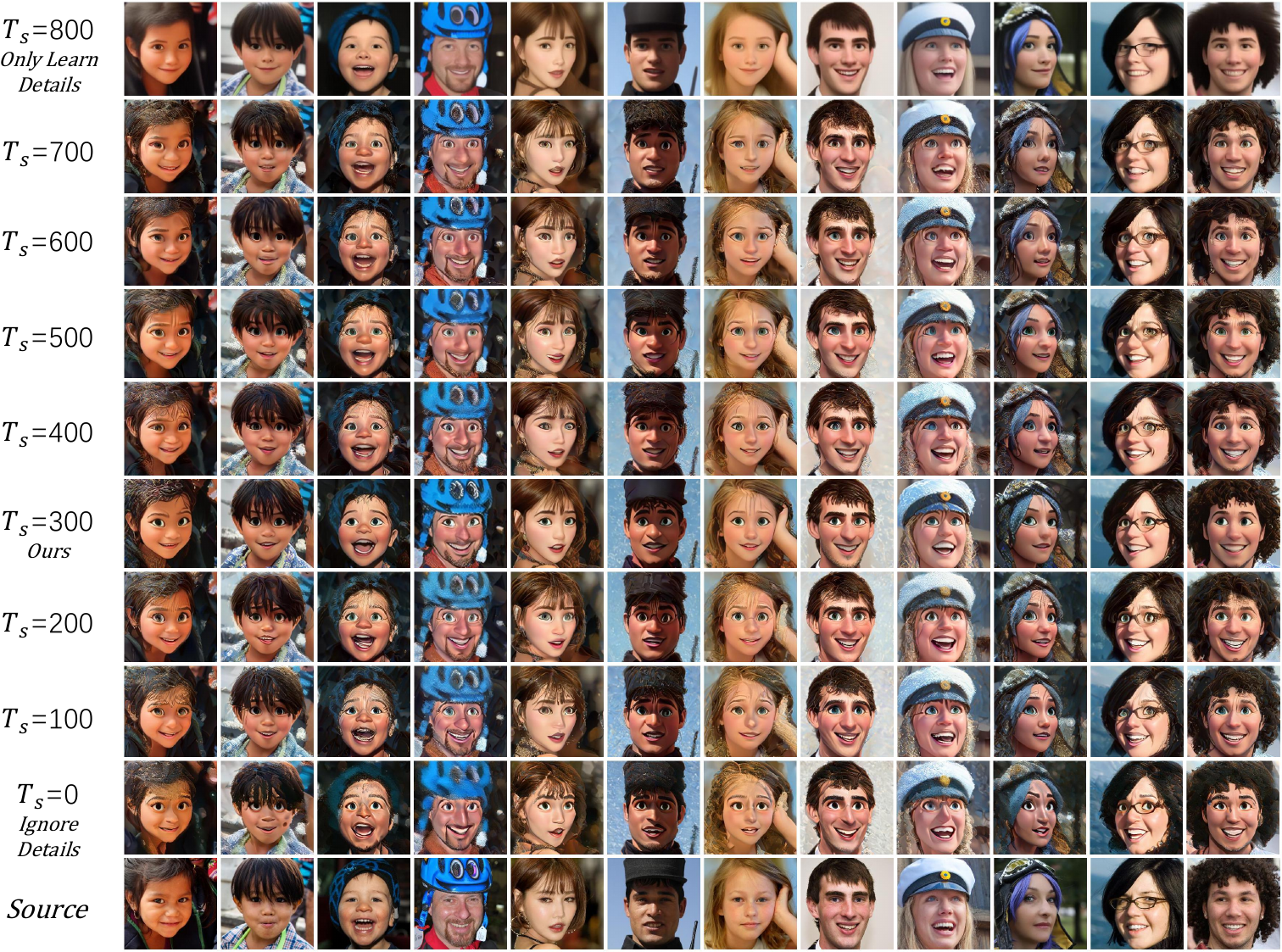}
\caption{\textbf{Ablation study on $T_s$, the phasic factor in $m(t)$ on FFHQ $\to$ Cartoon~\cite{pinkney2020resolution}.} When $T_s=800$, the model only learns the target-domain details, ignoring the global style. When $T_s=0$, the model learns the style and content information in the whole process, which leads to failed style transfer at $t$-small, causing an unstable training which generates artifacts and rough details in the output images.}
\label{fig:ablation on T_s}
\end{figure*}
\section{Comparison with SDEdit}
\label{Sec: quantitative comparison}

SDEdit~\cite{meng2021sdedit} is a model that maintains the content information during domain adaption by adding a $t$-step noise into a source image and denoise it. In comparison, our model utilize phasic content fusion (PCF) module to keep the content information. Different from SDEdit which only keeps the content information in the noised image and has no further contents injected during the denoising stage, our PCF constantly fuses the images in the denosing process with the features from source image using a three-layer convolution network, thereby aiding our model in autonomously acquiring content information from the original images. To further validate the effectiveness of our PCF, we compare our PCF with SDEdit in content preservation and generation quality.

In addition to the Inception Score (IS), Structural Consistency Score (SCS), and Intra-cluster pairwise LPIPS distance (IC-LPIPS) metrics employed in the main paper, we also incorporate the Fréchet Inception Distance (FID)~\cite{heusel2017gans} to measure the similarity between the features of the source data and the generated data according to their mean values and covariance. A lower FID suggests generated data is similar to source data with high diversity and realism.

In order to facilitate a more effective comparison between PCF and SDEdit, we refrain from utilizing the ICSG module for contour preservation (PCF only) and substitute our PCF by SDEdit with different noising steps. We compute the FID, IS, IC-LPIPS and SCS scores in Tab.~\ref{tab:comparison with SDEdit}, where the noising step of SDEdit ranges from 400 to 600 (the recommended parameter in its paper). It can be seen that our PCF outperforms SDEdit in terms of generation quality and diversity.



\section{Ablation Study}

\label{Sec:ablation study}
\subsection{Ablation Study on The Phasic Factor $T_s$}
The phasic factor $T_s$ in $m(t)$ is an important parameter that influences the generated results. A large $T_s$ leads to the failure in style transfer since there are only $M-T_s$ steps to transfer the style. Similarly, a small $T_s$ leads to failure in capturing target-domain details, causing rough details in the generated images. Moreover, when $t$ is too small, the failure in style transfer also leads to an unstable training process which generates artifacts in the output images. To validate this, we conducted an ablation study on the phasic factor $T_s$ and show the results in Fig. \ref{fig:ablation on T_s}.

\begin{table}[t]
\vspace{-0.15in}
\small
\centering
\setlength{\abovecaptionskip}{2pt}
\setlength{\belowcaptionskip}{-0.2cm}
\setlength\tabcolsep{3pt}
\renewcommand{\arraystretch}{1.2}

\scalebox{0.8}{
\begin{tabular}{c|ccccccccc}
\toprule
$T_s$&0&100&200&300&400&500&600&700&800 \\\hline
FID&136.51&137.00&125.46&\textbf{119.65}&123.75&145.62&157.11&161.40&155.93\\
\bottomrule
\end{tabular}}
\label{tab:ablation on T_s}
\caption{\hut{\textbf{Ablation study on $T_s$}, the phasic factor in $m(t)$ on FFHQ $\to$ Cartoon~\cite{pinkney2020resolution}. We evaluate the FID between the generated images and the cartoon dataset. It can be seen that when $T_s$ ranges from 200 to 400, the FID scores are similar, indicating that a wide range of $T_s$ result in a good performance. }}
\label{tab: quantitative ablation on T_s}
\end{table}
\begin{table}[t]
\centering
\renewcommand{\arraystretch}{1.2}
\resizebox{1.0\linewidth}{!}{
\begin{tabular}{c|cccc}
\toprule
Model& Ours & IDC loss&RSSA loss &NADA loss\\ \midrule
Sketches&\textbf{47.42}&146.32&125.77&88.76\\
Cartoon&\textbf{119.65}&180.28&171.99&144.57\\
\bottomrule
\end{tabular}}
\caption{\textbf{Quantitative comparison between our DDC loss and the losses in IDC, RSSA and StyleGAN-NADA.}}
\label{tab:more quantitative comparison on DDC loss}
\end{table}
\begin{figure*}[t!]
\centering
\includegraphics[width=0.98\textwidth]{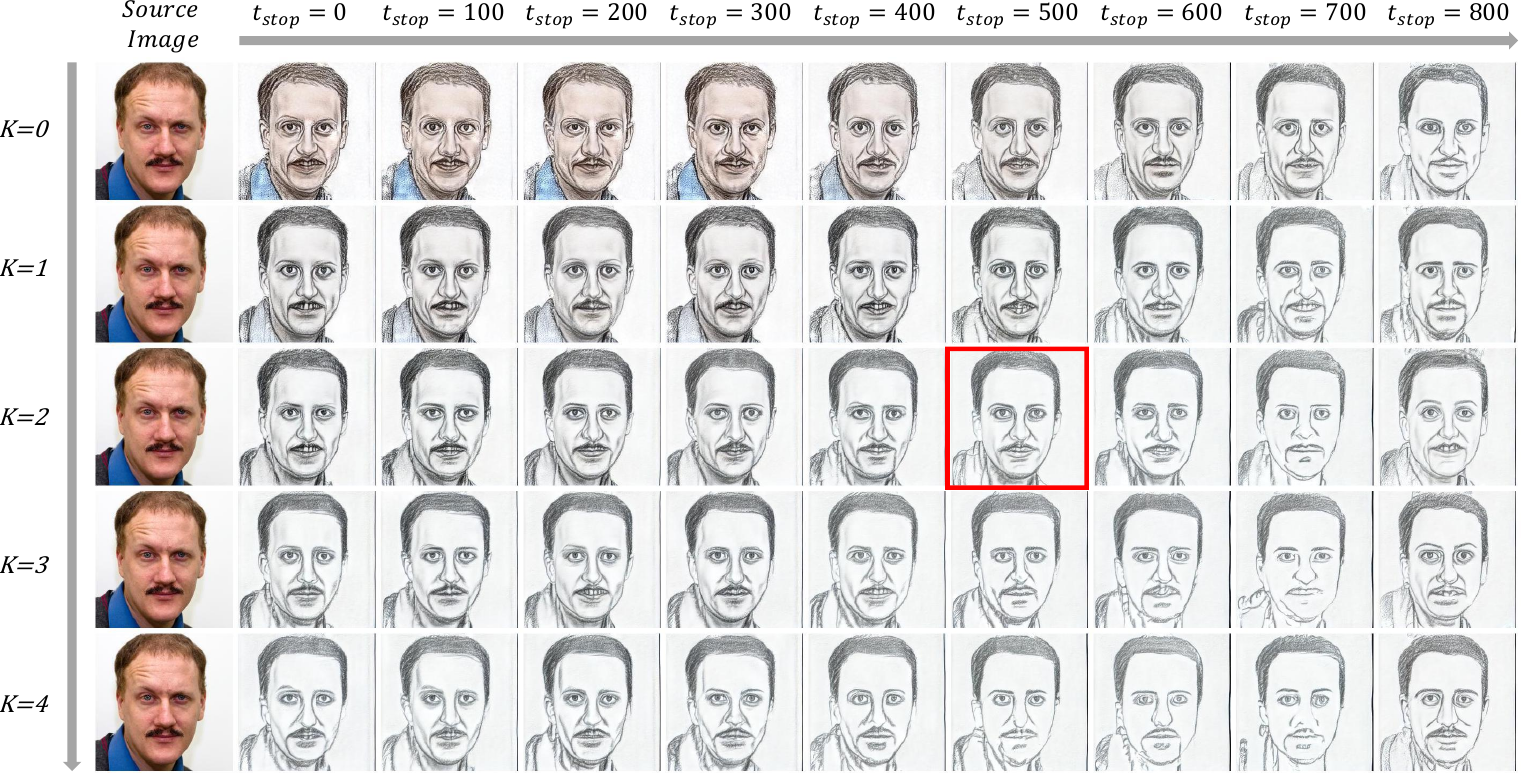}
\caption{\textbf{Ablation Study on the stop step $t_s$ and the repeating factor $K$ in style enhancement module} with the filtering factor $N=8$. As $t_{stop}$ or $K$ grows, the generated image shares less contents with the source image. When $t$ and $K$ is small, the model cannot eliminate the influence of the source image, i.e., generating wrong color in the output image.}
\label{fig:ablation on SE-K}
\vspace{-0.05in}
\end{figure*}
\begin{figure*}[t!]
\centering
\includegraphics[width=0.98\textwidth]{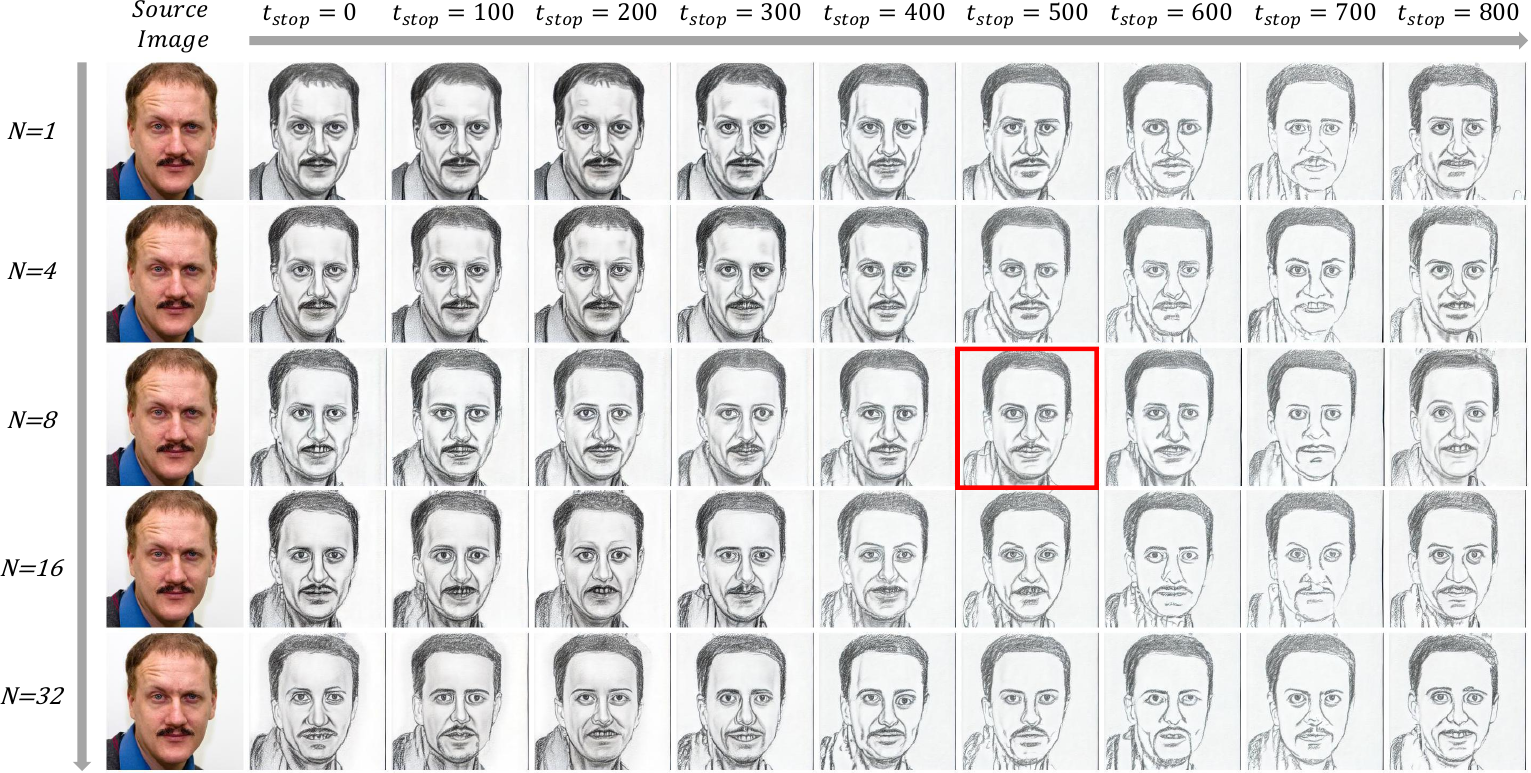}
\caption{\textbf{Ablation Study on the stop step $t_s$ and the filtering factor $N$} with the repeating factor $K=2$. As $t_{stop}$ or $N$ grows, the generated image shares less contents with the source image.}
\label{fig:ablation on SE-N}
\vspace{-0.05in}
\end{figure*}
It can be seen that when $T_s=800$, the model only transfers the local details in the target domain, ignoring the global style. When $T_s=0$, the model suffers from generating many artifacts, and the generated images all seem to be rough. Thus, we chose $T_s=300$ as our default setting, which balances both style transfer and detail capturing.

Moreover, we also compute the FID scores between the cartoon dataset and our generated data with different $T_s$. The results are shown in Tab.~\ref{tab: quantitative ablation on T_s}. It can be seen that when $T_s$ ranges from 200 to 400, the FID scores are similar, indicating that a wide range of $T_s$ result in a good performance.

\subsection{Ablation Study on ICSG}
\label{sec:Ablation Study on ICSG}
Our iterative cross-domain structure guidance strategy (ICSG) comprises three key parameters: the repeating factor $K$ of the style enhancement module, the filtering factor $N$, and the stop step $t_{stop}$.  To demonstrate a clear comparison among different parameter values, we conducted an experiment on the FFHQ $\to$ Cartoon task. 

Firstly, we investigate the influence of the repeating factor $K$ and stop step $t_{stop}$ on the output of ICSG, as shown in Fig.~\ref{fig:ablation on SE-K} with $N=8$. Next, we examine the impact of the filtering factor $N$ and stop step $t_{stop}$ on the output of ICSG, as illustrated in Fig.~\ref{fig:ablation on SE-N} with $K=2$. (It should be noted that the default setting in our method is $K=2$, $N=8$, and $t_{stop}=500$ here)
We observed that as $K$, $t_{stop}$, or $N$ increases, the model captures more style in the target domain but loses more content information and local structures. When both $K$ and $t_{stop}$ are small, the model cannot effectively eliminate the influence of the source image in terms of original color and texture. In summary, a bigger $K$ and $t_{stop}$ enhance the stylization effect and a smaller $K$, $t_{stop}$ and $N$ keep more content information. In general, a wide range of parameter values near our default setting ($K=2$, $N=8$, and $t_{stop}=500$) yield favorable outcomes as illustrated in  Fig.~\ref{fig:ablation on SE-K} and \ref{fig:ablation on SE-N}, indicating that our model is not too sensitive to the parameters.



\begin{figure}[t]
\centering
\includegraphics[width=0.45\textwidth]{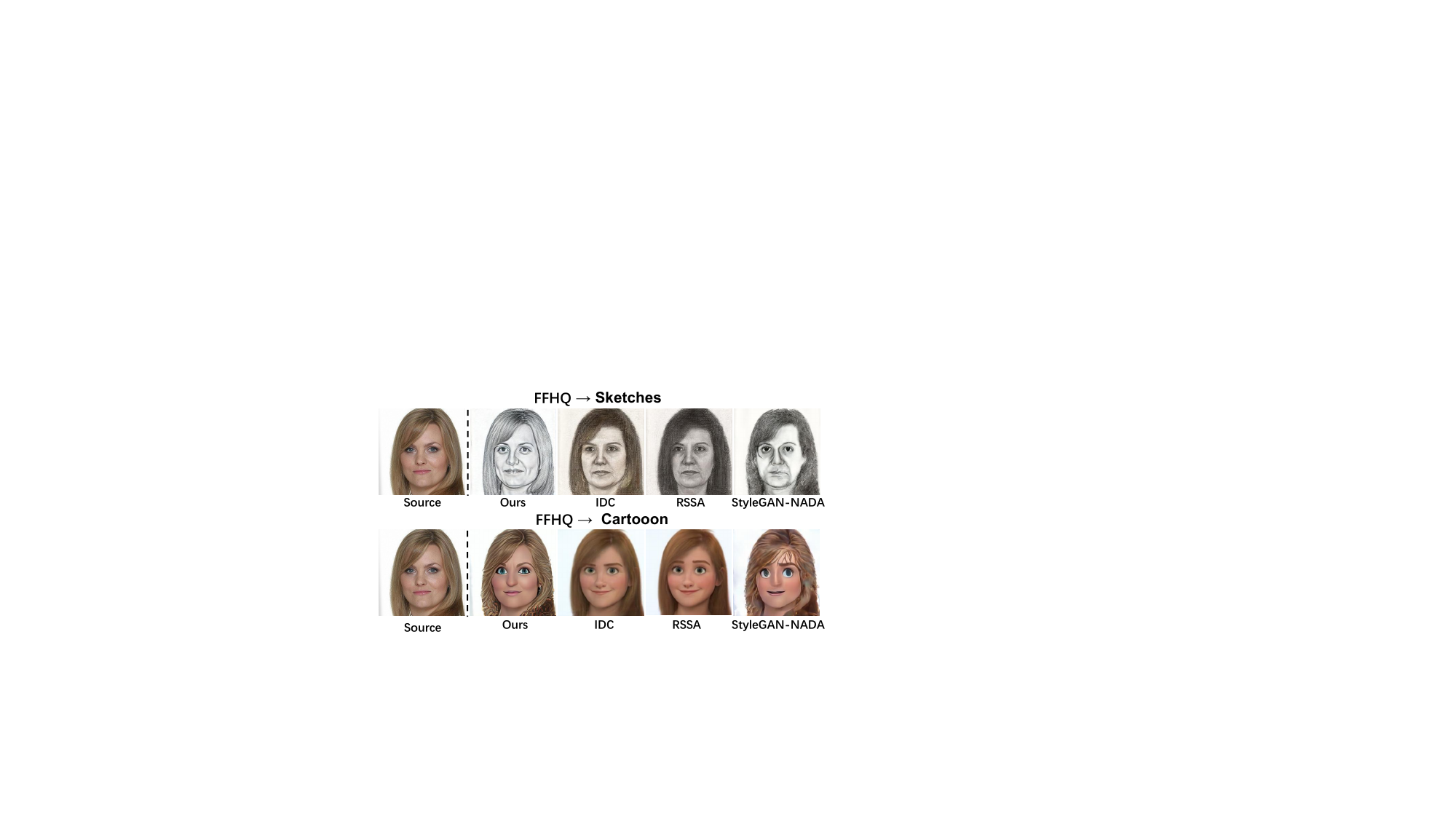}
\caption{\textbf{Qualitative comparison results between our DDC loss and the losses in IDC, RSSA and StyleGAN-NADA.}}
\label{fig:more qualitative comparison on DDC loss.}
\end{figure}

\section{More Ablation Study on DDC Loss}
We compare our DDC loss with the losses in IDC~\cite{ojha2021few}, RSSA~\cite{zhu2022few} and StyleGAN-NADA~\cite{gal2022stylegannada}. For a fair comparison, we exclusively substitute the DDC loss in our model with their losses, and keep the other conditions unchanged. The comparison results are shown in Tab.~\ref{tab:more quantitative comparison on DDC loss} and Fig.~\ref{fig:more qualitative comparison on DDC loss.}. It can be seen that our DDC loss outperforms the other distribution-consistency losses in diffusion-based few-shot domain adaption.

\section{Details of Iterative Cross-domain Structure Guidance (ICSG)}
\label{Sec:ICSG}
\subsection{Far More Than Few-shot Image Translation}
In our main paper, we introduce a novel iterative cross-domain structure guidance strategy (ICSG) for image sampling, which helps to retain structural information. The proposed ICSG is not limited to few-shot image translation tasks but can be applied to any image-to-image translation task on any source and target domains. In this section, we aim to demonstrate the effectiveness of ICSG and show more experiments on image-to-image translation.

\subsection{Derivation of ICSG}
In this section, we provide theoretical proof derivations to explain why our method is effective. For the sake of convenience, we define the following notations:
\begin{definition}
We define the denoising process $\Theta_t$ as:
\[ \Theta_t: \mathbf{R}^D \longrightarrow \mathbf{R}^D \]
\begin{equation}
\qquad x_t \mapsto x_{t-1} \nonumber
\end{equation} 
\[
\Theta_t(x_t)=\frac{1}{\sqrt{\alpha_t}}(x_t-\frac{1-\alpha_t }{\sqrt{1-\bar{\alpha}_t}}\epsilon_{\theta})+\sigma_t z \sim p_{\theta}(x_{t-1}|x_t)~,
\]
where $D$ is the dimension of the image, and $z$ is a random variable from standard normal distribution.
\end{definition}
\begin{definition}
We define the forward process $ \Phi_t $ as:
\[\Phi_t:\mathbf{R}^D \longrightarrow \mathbf{R}^D\]
\begin{equation}
 \qquad x_0 \mapsto x_t \nonumber
\end{equation}
\[
\Phi(x_0)=x_t(x_0,\epsilon)=\sqrt{\bar{\alpha}_t}x_0+\sqrt{1-\bar{\alpha}_t}\epsilon \sim q(x_t|x_0)~.
\]
\end{definition}
\begin{definition}
We define the backward process $ \Psi_t $ as:
\[\Psi_t:\mathbf{R}^D \longrightarrow \mathbf{R}^D\]
\begin{equation}
 \qquad x_t \mapsto \hat{x_0} \nonumber
\end{equation} 
\[
\Psi_t(x_t)=\frac{1}{\sqrt{\bar{\alpha}_t}}(x_t-\sqrt{1-\bar{\alpha}_t}\epsilon_{\theta}) \sim p_{\theta}(\hat{x_0}|x_t) .
\]
\end{definition}

Here, $x_0$ denotes the output image in the target domain, and $y_0$ denotes the source-domain image with the target structure $\phi_N(y_0)$. Our ICSG can be defined as follows:

\begin{small}
    $$ICSG(x_{t-1}|x_t,y_0)=\Theta_t(x_t)+\phi_N(SE(y_0))-\phi_N(\Theta_t(x_t))$$
    $$where\, SE(y_0)=\Theta_t \circ (\Phi_t \circ \Psi_t)^n \circ \Phi_t(y_0)~,$$ 
\end{small}
where $ (\Phi_t \circ \Psi_t)^n$ is the style enhancement module.
We have the following theorem:
\begin{theorem}
With our ICSG, the generated image $x_0$ shares the same structure with the reference image $y_0$.
\begin{equation}
\mathbf{E}_{x_0 \sim p_{t(data)}}(\phi_N(x_0))=\mathbf{E}_{y_0 \sim p_{s(data)}}(\phi_N(y_0)) ~,
\end{equation}
where $p_{t(data)}$ is the target data distribution, and $p_{s(data)}$ is the source data distribution. This indicates that the structure of the output image $x_0$ is the same as that of the reference image $y_0$.

\end{theorem}

\noindent\textbf{Proof 1.}
When t is small, $\Psi_{t-1}(x_{t-1})$ is very close to $x_0$ and $\phi_N(x)$ further blurs them. Thus, we can approximate $\phi_N(x_0)$ as $\phi_N(\Psi_{t-1}(x_{t-1}))$.
\begin{equation}
\begin{aligned}
&\mathbf{E}_{x_0 \sim p_{t(data)}}(\phi_N(x_0))\nonumber \\
& \approx \mathbf{E}(\phi_N(\Psi_{t-1}(x_{t-1}))) \\
& =\mathbf{E}(\phi_N(\Psi_{t-1}(\Theta_t(x_t)+\phi_N(SE(y_0))-\phi_N(\Theta_t(x_t)))))\\
& = \mathbf{E}(\phi_N(\Psi_{t-1}(\Theta_t(x_t)-\phi_N(\Theta_t(x_t))))) \cdots Part I\\
& +\mathbf{E}(\phi_N(\Psi_{t-1} ( \phi_N(SE(y_0))))~. \cdots Part II
\end{aligned}
\end{equation}

Regarding $Part I$, we can utilize the linear properties of $\phi_N$ and $\Psi_{t-1}$, which yields:
\begin{equation}
\begin{aligned} 
 Part I= & \mathbf{E}(\phi_N(\Psi_{t-1} \circ \Theta_t(x_t)-\Psi_{t-1}( \phi_{N_s}(\Theta_t(x_t))))) \nonumber \\
 = & \phi_N(\mathbf{E}(\Psi_{t-1}(\frac{1}{\sqrt{\alpha_t}}(x_t-\frac{1-\alpha_t }{\sqrt{1-\bar{\alpha}_t}}\epsilon_{\theta})+\sigma_t z ))\\
 -&\mathbf{E}(\Psi_{t-1}(\phi_{N_s}(\Theta_t(x_t)))))\\
 = & \phi_N(\mathbf{E}(\frac{1}{\sqrt{\alpha_t}}\Psi_{t-1}(x_t)-\frac{1}{\sqrt{\alpha_t}} \Psi_{t-1}(\phi_{N_s}(x_t))))\\
 =& \frac{1}{\sqrt{\alpha_t}}\phi_N(\mathbf{E}(\Psi_{t-1}(x_t)-\Psi_{t-1}(\phi_{N_s}(x_t))))\\
\approx & 0~,
\end{aligned}
\end{equation}
note that $\mathbf{E}(\epsilon_{\theta})=0$ and $\mathbf{E}(z)=0$.
Furthermore, the last approximation holds, given that $\alpha_t$ is close to 1 when t is small and our filtering factor $N$ is not large (we have set $N$ to be 8).

Regarding Part II, when t is small, $\Psi_{t-1}(y_{t-1})$ is in close proximity to $y_0$, and $\phi_N(x)$ further blurs them. Therefore, we have:
\begin{equation}
\Psi_{t-1} ( \phi_N(SE(y_0))=\hat{y_0} \nonumber
\end{equation} 
\begin{equation}
Part II= \mathbf{E}(\phi_N(\hat{y_0} )) \approx \mathbf{E}(\phi_N(y_0 ))~. \nonumber
\end{equation}

Thus, we have demonstrated that:
\begin{equation}
\mathbf{E}_{x_0 \sim p_{t(data)}}(\phi_N(x_0))=\mathbf{E}_{y_0 \sim p_{s(data)}}(\phi_N(y_0)) ~.
\end{equation}

\subsection{Algorithm}
The process of our ICSG can be summarized in Alg.~\ref{alg:ICSG}.

\begin{algorithm}
	\renewcommand{\algorithmicrequire}{\textbf{Input:}}
	\renewcommand{\algorithmicensure}{\textbf{Output:}}
        \algorithmicrequire{ Source image $x$ and reference image $y_0$}
        
        \algorithmicensure{ Generated image $x_0$}
	\caption{ICSG for image-to-image translation}
	\begin{algorithmic}[1]
        \label{alg:ICSG}
            \STATE $x_M  \sim q(x_M|x) $ 
            \FOR{$t=M,...,1$} 
                \STATE $z \sim \mathcal{N}(0,I)$
                    \IF{$t \geq t_{stop}$}
                        \STATE $y_t  \sim q(y_t|y_0)$
                        \FOR{$i=1,...,K$} 
                            \STATE $\hat{y_0}  \sim p_{\theta}(\hat{y_0}|y_t)$
                            \STATE $\hat{y_t}  \sim q(\hat{y_t}|y_0)  \qquad  \triangleright Style\,Enhancement$
                            \STATE $y_t \leftarrow \hat{y_t}$
                        \ENDFOR
                        \STATE $y_{t-1}'  \sim p_{\theta}(y_{t-1}|y_t)$
                        \STATE $x_{t-1}'  \sim p_{\theta}(x_{t-1}| x_t)$
                        \STATE $x_{t-1} \leftarrow x_{t-1}'+\phi_N(y_{t-1}')-\phi_N(x_{t-1}')$
                    \ENDIF
                \STATE $x_{t-1} \sim p_{\theta}(x_{t-1}|x_t)$
            \ENDFOR
            \RETURN $x_0$
	\end{algorithmic}  
\end{algorithm}
        
\subsection{Comparison on Image-to-image Translation}
\begin{figure*}[t]
\centering
\includegraphics[width=0.8\textwidth]{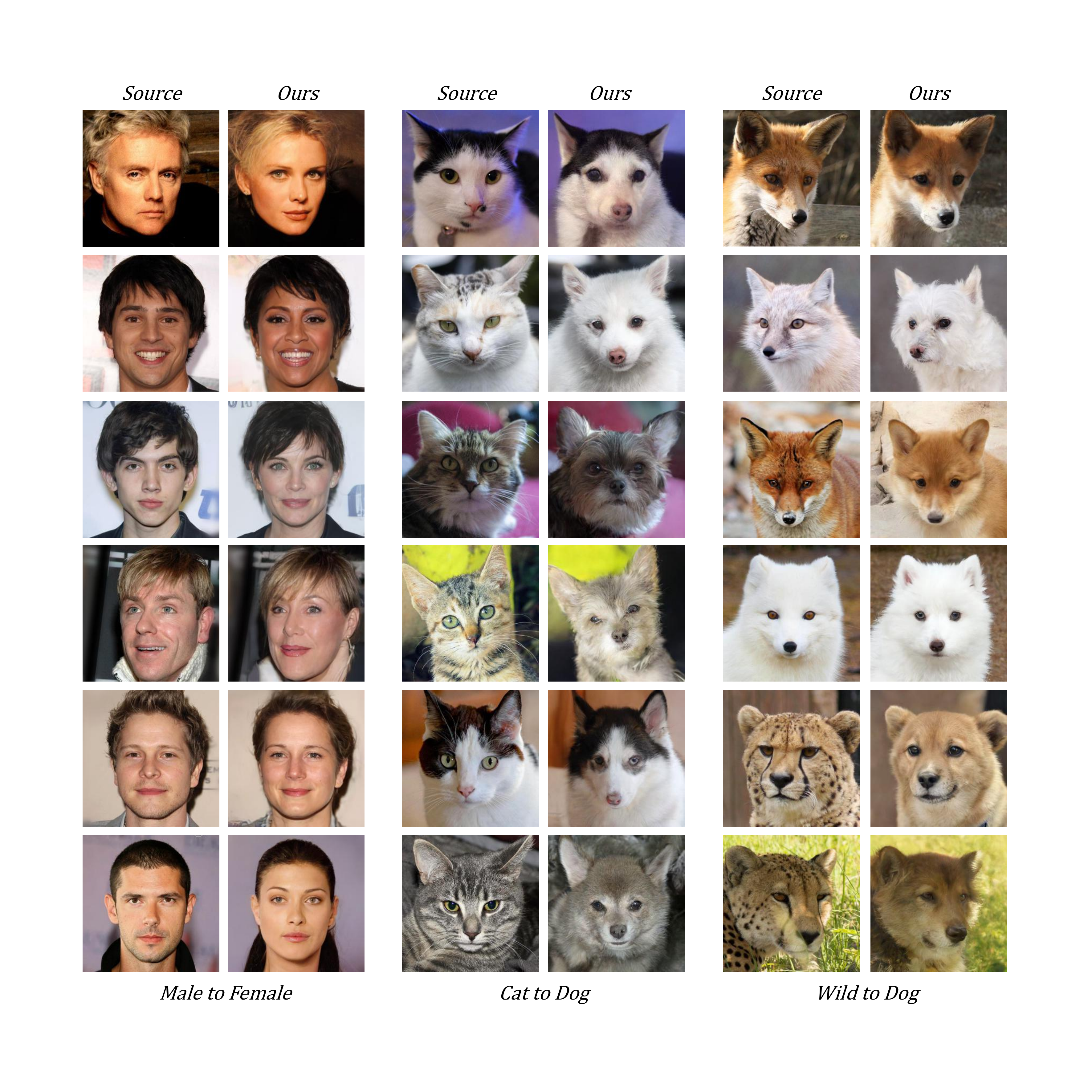}
\caption{Our ICSG results on male-to-female, cat-to-dog and wild-to-dog image-to-image translation. The generated results achieve  good performance in both structure preservation and domain translation.}
\label{fig:our results}
\vspace{-0.05in}
\end{figure*}
\begin{figure*}[t]
\centering
\includegraphics[width=0.98\textwidth]{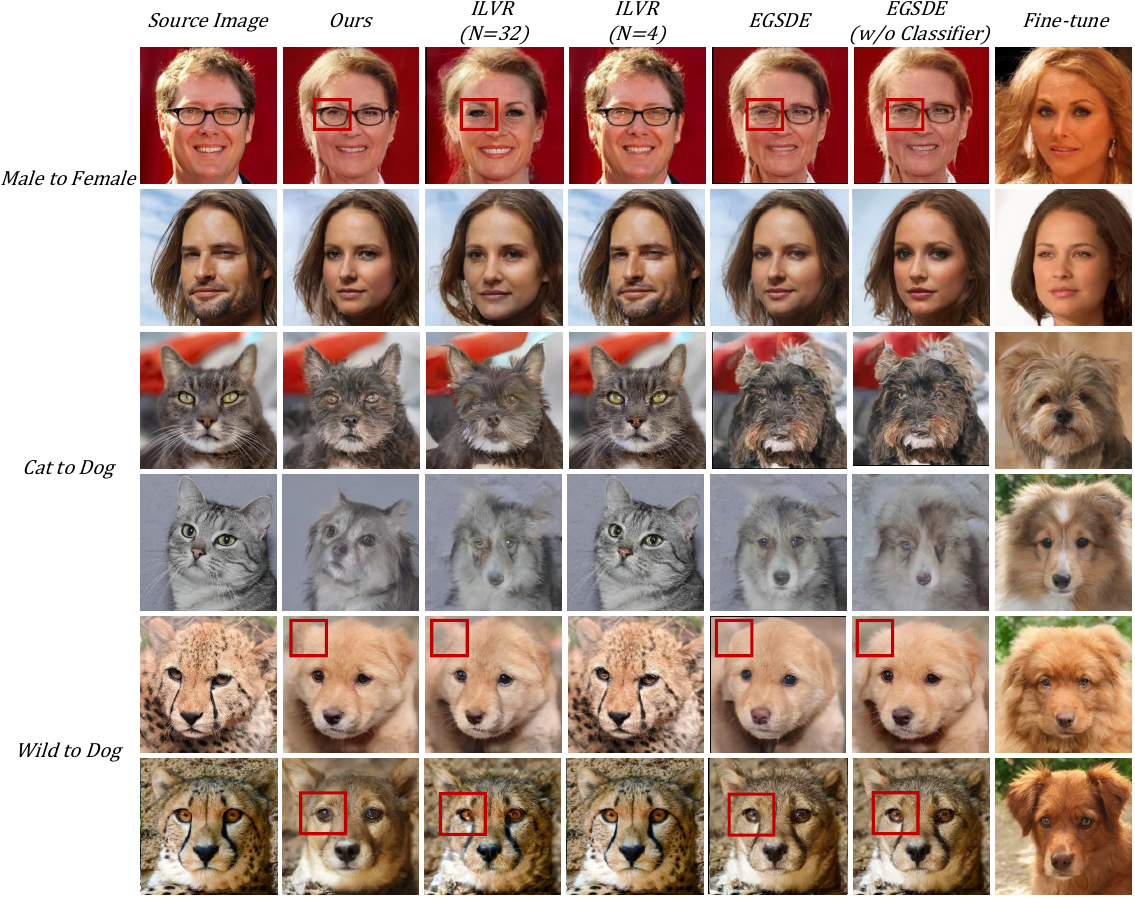}
\caption{Comparison between our method and existing methods based on diffusion model in domain transfer. From lest to right, the images are: images from source domain, Our model, ILVR with filtering factor $N=32$, ILVR with filtering factor $N=4$, EGSDE, EGSDE without classifier and the fine-tuned model.}
\label{fig:comaprison with EGSDE and ILVR}
\vspace{-0.05in}
\end{figure*}
In this section, we present additional experimental results on the image-to-image translation task. Specifically, we compare our proposed method with two existing diffusion-based image-to-image translation methods, namely EGSDE~\cite{zhao2022egsde} and ILVR~\cite{choi2021ilvr}, on cat-to-dog~\cite{choi2020stargan}, male-to-female~\cite{karras2018progressive} and wild-to-dog~\cite{choi2020stargan} image-to-image translation task. To ensure a fair comparison, we use the pretrained diffusion model provided in the source code of EGSDE for all experiments and set the default parameters for both EGSDE and ILVR.

Fig.~\ref{fig:our results} shows the image translation results obtained by our proposed ICSG method. It can be seem that our model performs well in terms of both domain translation and structure preservation.

To provide a more comprehensive comparison, we also compare our method with state-of-the-art methods EGSDE~\cite{zhao2022egsde} and ILVR~\cite{choi2021ilvr}. We use the pretrained diffusion model as a baseline, which adds 700-step noise to $x_0$ and denoises it. For ILVR, we use filtering factors $N$ of 4 and 32, which were effective in their paper. Moreover, since none of our method, ILVR, nor the fine-tuned model relies on additional classifiers, we conduct experiments on EGSDE with and without a classifier separately. The comparison results are shown in Fig.~\ref{fig:comaprison with EGSDE and ILVR}. The fine-tuned model loses much of the structural information after denoising. As for ILVR, when the filtering factor $N=32$, the translated images lose much structural information as well, and when $N=4$, it tends to reconstruct the source images. There is no distinct difference between the two EGSDE results with and without a classifier. However, both of them lose some structural information in the generated images. In contrast, our model achieves good performance in both structure preservation and image translation.

\section{Analysis on Directional Distribution Consistency Loss}
\label{Sec:DDC loss}
Our directional distribution consistency loss explicitly constrains the centrality consistency between the generated distribution and the target distribution, while preserving the structural consistency of the generated distribution and the original distribution. In this section, we provide a theoretical analysis to demonstrate that the prior loss functions in the existing few-shot image generation tasks share similar goals with our approach, but they suffer from the distribution rotation problem, which can cause unstable training and low training efficiency.

IDC \cite{ojha2021few} proposes a cross-domain distance consistency loss that can maintain the structure of the generated distribution and prevent overfitting. Based on this, RSSA~\cite{zhu2022few} further designs a cross-domain spatial structural consistency loss that can solve the drift problem of the generated samples in the target domain. However, both methods lack a deep analysis of the loss functions and suffer from distribution rotation during the training process. For the sake of convenience in our derivation, we use the loss function of IDC as an example for demonstration (since RSSA only addresses the issue of distribution drift, its proof follows a similar process).



To make our analysis clearer, we denote $x$ and $y$ as the latent variables, $P_S(x)$, $P_T(x)$ and $P_G(x)$ as the source, target and generated distribution, $S(x)$, $T(x)$ and $G(x)$ as the corresponding images of the latent variable $x$ in the source, target and generated distribution, and $C(\cdot)$ as the distribution center.


To minimize the few-shot loss in IDC, it satisfies:
\begin{small}
\begin{equation}
    \begin{aligned}
        cos(S(x_i),S(x_j))=cos(G(x_i),G_(x_j)) \quad \forall x_i,x_j\in P_S(x)~.
        \label{eq:few-shot loss}
    \end{aligned}
\end{equation}
\end{small}

Thus, the structure of the generated domain is fixed. The generated distribution $P_G(x)$ can only rotate or move along the axis that crosses the origin and distribution center $C(G)$ (note that when moving the distribution along the axis, the scale of the distribution also varies.). So, the major concern is to determine where the center of the generated distribution $C(G)$ is located. Based on our analysis, we present the following theorem:

\begin{theorem}
The center of the generated distribution $C(G)$ coincides with
that of the target distribution $C(T)$ with the adversarial loss and few-shot loss:
\begin{small}
\begin{equation}
\int_xG(x)P_G(x)dx=\int_xT(x)P_T(x)dx~.
\end{equation}
\end{small}
\end{theorem}
\textbf{Proof 2} The adversarial loss in GANs\cite{goodfellow2020generative} is to find a generator $G$ that satisfies $P_G(x)=P_T(x)$, which can be transferred into: $G(x)=T(x)$ , $\forall x \sim P_T(x)$. In high-dimensional space, we can employ cosine distance to measure the similarity. Then, we rewrite the goal of the adversarial loss as:
\begin{small}
\begin{equation}
\begin{aligned}
G=\mathop{argmin}\limits_{G} \mathbf{E}_{x \sim P_T(x)}|cos(G(x),T(x))-1|~. \nonumber
\end{aligned}
\end{equation}
\end{small}

According to Eq.(\ref{eq:few-shot loss}), we can also rewrite the goal of the few-shot loss functions as:
\begin{small}
\begin{equation}
\begin{aligned}
G=&\mathop{argmin}\limits_{G} \frac{1}{2}\mathbf{E}_{x,y \sim P_{S}(x)}|cos(S(x),S(y))-cos(G(x),G(y))|~. \nonumber
\end{aligned}
\label{eq:DDC-structure}
\end{equation}
\end{small}

Combining both the adversarial and few-shot loss together, we have the final optimization goal:
\begin{small}
\begin{equation}
\begin{aligned}
G&=\mathop{argmin}\limits_{G}  \mathbf{E}_{x \sim P_T(x)}|cos(G(x),T(x))-1|+ \\
&\frac{1}{2}\mathbf{E}_{x,y \sim P_{S}(x)}|cos(S(x),S(y))-cos(G(x),G(y))| ~.
\end{aligned}
\label{eq:DDC1}
\end{equation}
\end{small}

In high-dimensional space, any tow points has almost the same Euclidean distance, indicating the the modulus of the each vector are extremely close, denote the modulus as $\sqrt\lambda$. So, we transform Eq.(\ref{eq:DDC1}) into:
\begin{small}
\begin{equation}
    \begin{aligned}
        G&=\mathop{argmin}\limits_{G}  \int_x|G(x)T(x)-\lambda|P_T(x)dx+ \\
&\frac{1}{2}\int_x\int_y|S(x)S(y)-G(x)G(y)|P_G(x)P_G(y)dxdy~.
    \end{aligned}
    \label{eq:DDC2}
\end{equation}
\end{small}

Taking the gradient on $G$ in Eq.(\ref{eq:DDC2}) and with the symmetry property of $x$ and $y$, the optimal $G^*$ satisfies:
\begin{small}
\begin{equation}
\begin{aligned}
& \int_xT(x)P_T(x)dx-\int_x\int_yG(x)P_G(x)P_G(y)dxdy=0  \\
\iff& \int_xT(x)P_T(x)dx=\int_xG(x)P_G(x)dx\int_yP_G(y)dy \\
\iff&\int_xG^*(x)P_{G^*}(x)dx=\int_xT(x)P_T(x)dx~.
\end{aligned}
\label{DDC:3}
\end{equation}
\end{small}

The optimal generation distribution aligns with the center of the target distribution (as shown in Eq.(\ref{DDC:3})). Once the distribution center is fixed, the generated distribution cannot shift along the axis that passes through the origin and the center. Therefore, the scale of the generated distribution matches that of the source distribution. Unfortunately, this does not solve the issue of distribution rotation, which can result in an unstable and ineffective training process.

In contrast, our directional distribution consistency loss maintains the distribution center and structure explicitly and any rotation or shift of the generated distribution result in an increase of our loss function.

\end{document}